\pdfoutput=1
\documentclass{article}

\usepackage[final]{neurips_2025}

\usepackage[utf8]{inputenc} 
\usepackage[T1]{fontenc}    
\usepackage{hyperref}       
\usepackage{url}            
\usepackage{booktabs}       
\usepackage{amsfonts}       
\usepackage{nicefrac}       
\usepackage{microtype}      

\usepackage{graphicx}
\usepackage{booktabs} 
\usepackage{multirow}
\usepackage{bm}
\usepackage{amsmath}
\usepackage{amsthm}

\usepackage{caption}
\usepackage{float} 
\usepackage{amsmath,amssymb,amsfonts}
\usepackage{algorithmic}
\usepackage{textcomp}
\usepackage{bm}
\usepackage{amsfonts}
\usepackage[table]{xcolor}
\usepackage{diagbox}
\usepackage{booktabs}
\usepackage[normalem]{ulem}
\usepackage{multirow}
\usepackage{pifont}
\usepackage{enumitem}
\usepackage{booktabs}
\usepackage{amssymb}
\usepackage{wrapfig}
\usepackage{tikz}
\theoremstyle{plain}
\newtheorem{theorem}{Theorem}[section]

\theoremstyle{definition}

\newtheorem{assumption}[theorem]{Assumption}
\theoremstyle{remark}

\usepackage{amsmath}
\usepackage{amssymb}
\usepackage{mathtools}
\usepackage{amsthm}
\usepackage{colortbl}
\usepackage{tikz}      
\usepackage{colortbl}
\usepackage{float}
\usepackage{hyperref}
\usepackage{listings}
\usepackage[linesnumbered,ruled,vlined]{algorithm2e}

\usepackage{amsmath}
\usepackage{multirow}
\newcommand{\cmark}{\textcolor{green!50!black}{\ding{51}}}
\newcommand{\xmark}{\textcolor{red}{\ding{55}}}
\newcommand{\neutral}{\textcolor{blue!50!white}{\ding{108}}}
\usepackage{apptools}
\usepackage{caption}
\usepackage{chngcntr}

\title{Harmony in Divergence: Towards Fast, Accurate, and Memory-efficient Zeroth-order LLM Fine-tuning}

\author{%
Qitao Tan\textsuperscript{1} \quad
Jun Liu\textsuperscript{2} \quad
Zheng Zhan\textsuperscript{2} \quad
Caiwen Ding\textsuperscript{3} \\
\textbf{Yanzhi Wang}\textsuperscript{2} \quad
\textbf{Xiaolong Ma}\textsuperscript{4} \quad
\textbf{Jaewoo Lee}\textsuperscript{1} \quad
\textbf{Jin Lu}\textsuperscript{1} \quad
\textbf{Geng Yuan}\textsuperscript{1} \\
\textsuperscript{1}University of Georgia \quad
\textsuperscript{2}Northeastern University \\
\textsuperscript{3}University of Minnesota \quad
\textsuperscript{4}University of Arizona \\[0.5em]
}

\begin{document}

\maketitle

\begin{abstract}
Large language models (LLMs) excel across various tasks, but standard first-order (FO) fine-tuning demands considerable memory, significantly limiting real-world deployment. Recently, zeroth-order (ZO) optimization stood out as a promising memory-efficient training paradigm, avoiding backward passes and relying solely on forward passes for gradient estimation, making it attractive for resource-constrained scenarios. However, ZO method lags far behind FO method in both convergence speed and accuracy. To bridge the gap, we introduce a novel layer-wise divergence analysis that uncovers the distinct update pattern of FO and ZO optimization. Aiming to resemble the learning capacity of FO method from the findings, we propose \textbf{Di}vergence-driven \textbf{Z}eroth-\textbf{O}rder (\textbf{DiZO}) optimization. DiZO conducts divergence-driven layer adaptation by incorporating projections to ZO updates, generating diverse-magnitude updates precisely scaled to layer-wise individual optimization needs. Our results demonstrate that DiZO significantly reduces the needed iterations for convergence without sacrificing throughput, cutting training GPU hours by up to 48\% on various datasets. Moreover, DiZO consistently outperforms the representative ZO baselines in fine-tuning RoBERTa-large, OPT-series, and Llama-series on downstream tasks and, in some cases, even surpasses memory-intensive FO fine-tuning. Our code is released at \url{https://github.com/Skilteee/DiZO}.
\end{abstract}

\section{Introduction}

Fine-tuning large language models (LLMs) via backpropagation achieves strong performance across many NLP tasks~\citep{yang2019end, liu2019roberta,talmor2018commonsenseqa,chowdhery2023palm,zheng2020end}, but their large parameter counts create substantial memory burdens, limiting downstream applicability. Following neural scaling laws~\citep{hoffmann2022empirical,kaplan2020scaling}, next-generation LLMs grow rapidly, e.g., model sizes increase 410× every two years, far outpacing DRAM bandwidth (1.4×) and capacity (2×) growth. This imbalance leads to the \emph{memory wall}\citep{gholami2024ai}, a growing challenge especially for deployment on memory-limited devices\citep{zeng2024flightllm,chen2024understanding,tan2025perturbation}.


Zeroth-order (ZO) optimization has recently emerged as a memory-efficient approach for LLM fine-tuning, gaining growing attention~\citep{zhang2024revisiting,liu2024sparse,malladi2023fine,zhao2024second}. By relying solely on forward passes for gradient estimation, ZO eliminates the need for backpropagation and significantly reduces memory usage for activations, gradients, and optimizer states. As shown in~\cite{malladi2023fine}, ZO fine-tuning can reduce memory cost by up to 12$\times$. However, ZO still shows a notable \textbf{gap} in convergence speed and accuracy compared to first-order (FO) methods, as shown in  Table~\ref{table1}. Although ZO benefits from higher throughput due to its simpler computation, it requires over 10× more iterations to converge, ultimately increasing GPU time. Prior work often attributes this gap to noisy gradient estimates, without exploring other contributing factors~\citep{malladi2023fine, gautam2024variance, zhao2024second}.

\begin{wraptable}{r}{0.5\textwidth}
\centering
\small
\caption{Fine-tuning results on SST-2 datasets. Although ZO method shows advantages in memory saving, left behind FO method in terms of both accuracy and GPU hours.
}
\begin{tabular}{lcccc}
\toprule
Model                     & \textbf{Type} & \textbf{Acc.} & \textbf{Memory} & \textbf{\begin{tabular}[c]{@{}c@{}}GPU \\ Hours\end{tabular}} \\ \midrule
\multirow{2}{*}{RoBERTa}  & FO            & 91.9         & 9.2 GB                                                                    & 12.3\%                                                        \\
                          & ZO            & 90.5         & 4.5 GB                                                                  & 100.0\%                                                       \\ \midrule
\multirow{2}{*}{OPT-2.7B} & FO            & 94.2         & 45.4 GB                                                                   & 16.8\%                                                        \\
                          & ZO            & 90.0         & 6.8 GB                                                                  & 100.0\%                                                       \\ \bottomrule
\end{tabular}
\label{table1}
\end{wraptable}

To further uncover the fundamental causes of this gap, we begin by analyzing the distinct update patterns shown by FO and ZO methods during LLM fine-tuning. Interestingly, our analysis reveals a substantial difference in the magnitude of weight updates between layers. Specifically, FO methods benefit from fine-grained gradient estimation and enable \underline{diverse-magnitude updates} precisely scaled to the layer-wise individual optimization needs.
In contrast, ZO method tends to behave with \underline{uniform-magnitude updates} without considering layer-wise individual characteristics. This is attributed to the nature of ZO that relies on high-dimensional random search and leverages random perturbation for gradient estimation. Based on this, we conjecture that the compromised performance of ZO stems from its limited capability in achieving diverse-magnitude updates. This naturally raises the question: \emph{if we could enable ZO to achieve the desired diverse-magnitude updates, could we effectively achieve training acceleration and accuracy improvement?}

To validate our conjecture and fill the performance gap, we innovatively propose \textbf{Di}vergence-driven \textbf{Z}eroth-\textbf{O}rder optimization (\textbf{DiZO}), which performs divergence-driven layer adaptation via 
anchor-based learnable projections,
enabling principled adaptive updates that resemble FO methods.
Specifically, DiZO guides updates along geometrically constrained directions by learning target distances from an anchor point (e.g., the pre-trained model).
We also design a ZO-based method for projection learning that ensures the entire training process is memory-efficient.
We extensively evaluate DiZO on a range of tasks, including classification and generation, using models such as RoBERTa-large, the OPT series, and the Llama series. Results show that DiZO significantly reduces training iterations and cuts GPU hours by up to 48\% without sacrificing throughput. DiZO also integrates seamlessly with parameter-efficient tuning methods like LoRA~\citep{hu2021lora}, and consistently outperforms ZO baselines, sometimes even surpassing memory-intensive FO fine-tuning. Finally, we comprehensively analyze several potential alternatives and validate the necessity and effectiveness of our approach.

The summary of our contributions is as follows:
\begin{itemize}
    \item We introduce a novel layer-wise divergence analysis to uncover the fundamental differences in the updating patterns of FO and ZO methods.
    \item We introduce DiZO, a novel ZO method using divergence-driven layer adaptation, achieving a learning capacity closely resembling FO while maintaining the throughput benefit.
    \item DiZO consistently exceeds existing baselines in both accuracy and GPU hours, and it can be seamlessly integrated with LoRA for additional benefits. These advantages hold across diverse tasks and LLM architectures.
    \item We also provide comprehensive analysis and discussions on overheads, convergence guarantee, and potential alternatives, which further strengthen the efficiency, feasibility, and necessity of our proposed approach.
\end{itemize}

\begin{figure*}[ht]
\centering
\includegraphics[width=1.0\linewidth]{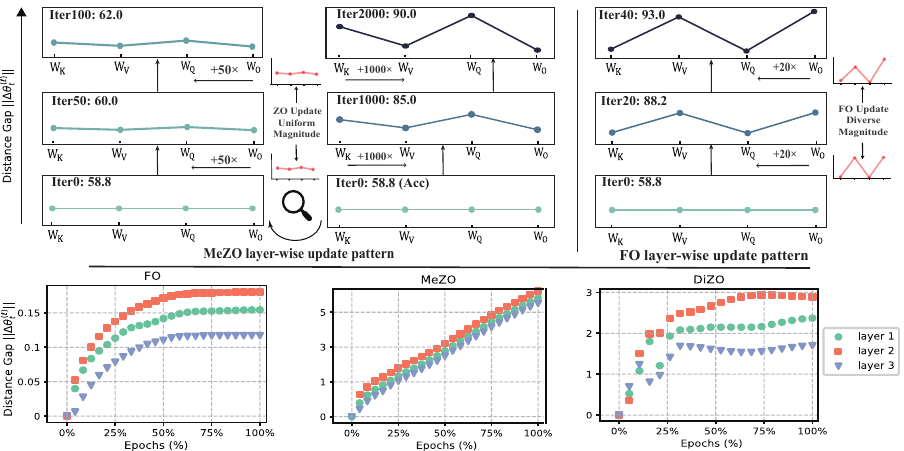}
\caption{Comparison of the training dynamics of ZO and FO methods. For the upper subfigure, $W_{K},W_{V},W_{Q},W_{O}$ indicate the corresponding weight matrix in the attention module.
}
\vspace{-10pt}
\label{zo_fo}
\end{figure*}
\section{Preliminaries and Pattern Analysis}

\subsection{Revisiting Zeroth-order Optimization}
Recently, ZO optimization has gained significant attention in machine learning~\citep{verma2023certified,dhurandhar2019model,wang2022zarts,gu2021efficient}.
Unlike conventional FO optimization, which calculates gradients via backpropagation, ZO optimization estimates gradients using only objective oracles via finite differences~\citep{chen2023deepzero, liu2018zeroth, ye2018hessian}. This property can be leveraged for LLM fine-tuning to alleviate the extensive memory costs. Specifically, as ZO only needs two forward passes to obtain the estimated gradients, it avoids computing and storing the most memory-consuming information needed in the conventional FO training, i.e., activations in the forward process, gradients in the backward process, and the optimizer state.

The core idea of ZO optimization is to estimate gradients by applying random perturbations to the weights and computing differences in the objective. For a mini-batch of data $\mathcal{B}$, sampled from a labeled dataset $\mathcal{D} = \{x_{i}, y_{i}\}_{i=1}^{|\mathcal{D}|}$, a model with parameters $\bm{\theta} \in \mathbb{R}^{d}$, where $d$ represents the dimension of the parameter space, and the corresponding loss function $\mathcal{L}(\bm{\theta}; \mathcal{B})$. The gradient is estimated as:
\begin{equation}
\label{equ:1}
    \nabla \mathcal{L}(\bm{\theta};\mathcal{B})=\frac{1}{q} \sum_{i=1}^q\left[\frac{\mathcal{L}\left(\bm{\theta}+\epsilon \bm{u}_{i};\mathcal{B}\right)-\mathcal{L}\left(\bm{\theta}-\epsilon \bm{u}_{i};\mathcal{B}\right)}{2 \epsilon} \bm{u}_{i}\right]
\end{equation}
where $\bm{u}_{i}\sim \mathcal{N}(0, \mathbf{I})$ is a random perturbation typically drawn from standard Gaussian distribution, $q$ is the number of queries, and $\epsilon > 0$ is a small perturbation scalar for smoothing.

Given the learning rate $\eta$ and the mini-batch data $\mathcal{B}_{t}$ at $t$-th iteration, once the estimated gradient $\nabla \mathcal{L} (\theta;\mathcal{B}_{t})$ is obtained,
then ZO-SGD updates the parameters with the following rule:
\begin{equation}
\label{equ:2}
    \bm{\theta}_{t+1} = \bm{\theta}_{t} - \eta \nabla \mathcal{L}(\bm{\theta}_{t};\mathcal{B}_{t})
\end{equation}

\subsection{Layer-wise Divergence Analysis}
\label{hypo}
ZO optimization applies uniform-magnitude updates across layers, with similar L2-norms per iteration (see Appendix~\ref{proof}). This uniformity may underlie its weaker performance. To explore how update divergence impacts convergence and accuracy, we analyze the training dynamics of ZO and FO methods.

\textbf{Analysis indicator.} To quantify the effect of updates, we adapt the layer-wise L2-norm distance gap between the weights of the pre-trained and the weights of fine-tuned model as an indicator.
The layer-wise L2-norm distance gap is defined as:
\begin{equation}
    \|\Delta \bm{\theta}_{t}^{(\ell)}\| = \|\bm{\theta}_{t}^{(\ell)} - \bm{\theta}_{0}^{(\ell)}\|_2
\end{equation}
where $t$ is $t$-th fine-tuning iteration, $\ell$ is $\ell$-th layer of the model, and $\bm{\theta}_{0}^{(\ell)}$ indicates the weights of $\ell$-th layer of pre-trained model.

\textbf{Analysis result.} Figure~\ref{zo_fo} compares the training dynamics of FO and ZO optimization methods, focusing on how each shapes the layer-wise distance gap between model parameters and the pre-trained initialization. Both methods show increasing divergence among layers over time, as illustrated by the twisting lines in the upper sub-figure, indicating that layers benefit from different levels of deviation from the pre-trained model. However, FO and ZO differ in how this divergence accumulates. FO uses fine-grained gradients to produce diverse-magnitude updates, quickly establishing meaningful distance gaps within a few iterations. In contrast, ZO performs random search with uniform-magnitude updates, requiring thousands of iterations to reach similar divergence. Additionally, FO-based methods show fast but converging growth in distance gaps, while ZO-based methods exhibit linear, unconstrained growth. This continued expansion under ZO may reflect a lack of effective constraint, contributing to accuracy drops in later stages and resulting in suboptimal performance.

\begin{algorithm}[t]
\small  
\SetAlgoLined
\SetNlSty{textbf}{}{}  
\caption{Divergence-diven ZO Optimization (DiZO)}
\label{algorithm1}
\textbf{Require:} parameter of $t$-th iteration $\bm{\theta_{t}}$ and pre-trained model $\bm{\theta_{0}}$, loss function $\mathcal{L}$, step budget $T$, perturbation scalar $\epsilon$, mini-batch data $\mathcal{B}_{t}$, learning rate $\eta$, projection at $t$-th iteration $\bm{\gamma_t}=\{\gamma_{t}^{i}\}_{i=1}^{L}$, projection update interval $\kappa$

\For{$t = 1$ \KwTo $T$}{
    $\nabla \mathcal{L} = \text{\texttt{GradEst}}(\bm{\theta_{t}}, \epsilon, \mathcal{B}_{t})$\;

    $\bm{\theta_{t}} = \bm{\theta_{t-1}} - \eta\nabla \mathcal{L}$\;

    \If{$t \bmod \kappa = 0$}{
        $\bm{\gamma}^{*}_{t} = \arg\min_{\bm{\gamma}_t} \mathcal{L}(\bm{\theta}_{0} + \frac{\bm{\gamma}_t}{\|\Delta \bm{\theta}_{t}\|}\Delta \bm{\theta}_{t};\mathcal{B}_{t})$\;

        $\bm{\theta}_{t} = \text{\texttt{ApplyProjection}}(\bm{\theta}_t,\bm{\theta}_0,\bm{\gamma}_t^{*})$\;
    }
}

\SetKwFunction{FGradEst}{GradEst}
\SetKwProg{Subroutine}{Subroutine}{:}{\KwRet}
\Subroutine{\FGradEst{$\bm{\theta}$, $\epsilon$, $\mathcal{B}$}}{
    \textbf{Sample:} $u_{1},\dots,u_{q} \backsim \mathcal{N}(0, \mathbf{I})$\;

    \textbf{Query:} $y_{i} = \mathcal{L}(\bm{\theta}+\epsilon u_{i};\mathcal{B}) - \mathcal{L}(\bm{\theta}-\epsilon u_{i};\mathcal{B})$\;

    \textbf{Estimator:} $\nabla \mathcal{L} = \frac{q}{2\epsilon}\sum_{i=1}^{q}y_{i}u_{i}$\;

    \Return $\nabla \mathcal{L}$\;
}

\SetKwFunction{FApplyProjection}{ApplyProjection}
\SetKwProg{Subroutine}{Subroutine}{:}{\KwRet}
\Subroutine{\FApplyProjection{$\bm{\theta}_t$, $\bm{\theta}_0$, $\bm{\gamma}_t$}}{
    \For{$\ell=1,2, \dots, L$}{
        
        \tcp{Project $l$-th layer}
        $\bm{\theta}_t^{(\ell)} = \bm{\theta}_{0}^{(\ell)} + \frac{\gamma^{(\ell)}_t}{\|\Delta \bm{\theta}_{t}^{(\ell)}\|}\Delta \bm{\theta}_{t}^{(\ell)}$\;
    }
    \Return $\bm{\theta}_t$\;
}
\end{algorithm}
\section{Methodology}

\subsection{Design of the Divergence-driven Layer Adaptation}
To provide layer-wise adaptive updates for ZO optimization, we propose applying anchor-based learnable projections to the updates of different layers.
The pseudocode for the method is shown in Algorithm~\ref{algorithm1}.

Specifically, we treat training iteration as a two-step process that iteratively updates the weights and the projection factor.
Our approach involves two key steps performed in an alternating manner. First, we perform vanilla ZO optimization as defined in Eq.~(\ref{equ:2}). Second, we identify the ideal projections for the weights and apply them, generating the projected weights. Formally, we define the ideal projection learning as solving the following minimization problem:
\begin{equation}
\label{equ:3}
\min_{\bm{\gamma}_t} \mathcal{L}(\bm{\theta}_{0} + \frac{\bm{\gamma}_t}{\|\Delta \bm{\theta}_t\|}\Delta \bm{\theta}_t;\mathcal{B}_{t})
\end{equation}
where $\bm{\gamma}_{t}=\{\gamma_{t}^{(\ell)}\}_{\ell=1}^{L}$ is a projection vector at $t$-th iteration, and $L$ is the number of layers. 
While searching for the ideal projection, we freeze the model weights and use the same mini-batch data $\mathcal{B}_t$ that is employed for the main ZO weight fine-tuning.


After finding the ideal projection for the $t$-th ZO step, we project the weights as:
\begin{equation}
\label{equ:4}
    \bm{\theta}_{t} = \bm{\theta}_{0} + \frac{\bm{\gamma}_t}{\|\Delta \bm{\theta}_t\|}\Delta \bm{\theta}_t
\end{equation}
where we get the new $\bm{\theta}_t$ after projection, and then we use the projected one for the following fine-tuning. 
When the value of $\bm{\gamma_{t}}$ is larger than $\|\Delta \bm{\theta_t}\|$, it enlarges the distance gap between the fine-tuned model and the pre-trained model,
 and vice versa.










\subsection{How to Learn the Projection?}

Although promising, finding the ideal projection (defined in Eq.~(\ref{equ:3})) remains challenging due to the high complexity of the objective. A straightforward solution is to also perform backpropagation for gradient computation and optimize the projection accordingly (FO-based method). For example, we use Adam optimizer to directly update $\bm{\gamma}_{t}$. The results are shown in Table~\ref{compare1}, which significantly reduces 67.7\% of the iterations and 58.5\% of GPU hours, and increases by 3.4\% in accuracy.

\begin{wraptable}{r}{0.6\textwidth}
\small
\centering
\vspace{-10pt}
\caption{Fine-tuning OPT-2.7B on SST-2 dataset.~\neutral: partial gradient-free; DiZO\textsuperscript{\textdagger}: learning projection by FO method;
}
\begin{tabular}{lcccc}
\toprule
Task Type                                                         & \textbf{\begin{tabular}[c]{@{}c@{}}Gradient\\ Free\end{tabular}} & \textbf{Acc.} & \textbf{\begin{tabular}[c]{@{}c@{}}\#Train \\ Iter.\end{tabular}} & \textbf{\begin{tabular}[c]{@{}c@{}}GPU \\ Hours\end{tabular}} \\ \midrule
MeZO                                                              & \cmark                                            & 90.0         & 100\%                                                             & 100\%                                                         \\
DiZO\textsuperscript{\textdagger} (w. FO) & \neutral                                          & 93.4         & 33.3\%                                                            & 41.5\%                                                        \\
FT                                                                & \xmark                                            & 94.2         & 9.3\%                                                             & 16.8\%                                                        \\ \bottomrule
\end{tabular}

\label{compare1}
\end{wraptable}


However, searching projection with the FO method makes DiZO only partially gradient-free. Specifically, while the model weights are updated via ZO, the per-layer projection parameter $\gamma_t^{(\ell)}$ is updated via FO, which still requires the backward pass and storing memory-intensive activation. The only memory saving, compared to the vanilla FO fine-tuning, is the optimizer state. As a result, relying on FO to find the ideal projection, though it achieves faster convergence speed and better accuracy in ZO optimization, offers limited overall benefit. It is worth noting that the peak memory usage during training of the FO-based DiZO is similar to that of low-rank adaptation (LoRA)~\citep{hu2021lora}.





\subsection{Projection Learning by Zeroth-order Optimization}
\label{ZO}

Our major goal is to find the ideal projection for adaptive updates while avoiding memory-intensive backpropagation. One potential promising solution is to also utilize the ZO method to update the projection. We estimate the gradient and update the projection as:
\begin{gather}
\label{equ:5}
    \nabla \widehat{\mathcal{L}}(\bm{\gamma}_{t};\bm{\theta}_{t})=\left[\frac{\widehat{\mathcal{L}}\left(\bm{\gamma}_{t}+\epsilon \bm{u};\bm{\theta}_{t}\right)-\widehat{\mathcal{L}}\left(\bm{\gamma}_{t}-\epsilon \bm{u};\bm{\theta}_{t}\right)}{2 \epsilon} \bm{u}\right] \\
    \bm{\gamma}_{t,j+1} = \bm{\gamma}_{t, j} - \eta \nabla \widehat{\mathcal{L}}(\bm{\gamma}_{t};\bm{\theta}_{t})
\end{gather}
where $\bm{u} \in \mathbb{R}^{L}$ is a random vector from $\mathcal{N}(0, \mathbf{I})$ and $\widehat{\mathcal{L}}= \mathcal{L}(\bm{\theta_{0}} + \frac{\bm{\gamma_{t}}}{\|\Delta\bm{\theta_t}\|}\Delta\bm{\theta_t};\mathcal{B}_t)$.


However, directly applying vanilla ZO optimization to the sub-task of projection learning yields limited improvement and can even cause failure, undermining the main fine-tuning process (see Appendix~\ref{ablation}). This failure stems from two key issues. First, projection values depend not only on $\bm{\gamma}_t$ but also on the distance gap $\|\Delta\bm{\theta}_t\|$. Ignoring this gap leads to uninformative updates and suboptimal solutions. Second, due to noisy ZO updates over a few iterations, projection magnitudes can become too small or too large. A small projection pulls the model too close to the pre-trained state, erasing progress, while a large one applies overly aggressive updates that destabilize training.


To address the above issues, two strategies are devised. \\
\textbf{\uline{Re-initialization.}} To introduce the distance gap $\|\Delta \bm{\theta}_{t}\|$ into the projection learning process, the initial value $\bm{\gamma}_{t,0}$ is reset to $\|\Delta \bm{\theta}_{t}\|$ each time the projection is optimized. This means that, initially, the projection magnitude $\frac{\bm{\gamma}_t}{\|\Delta \bm{\theta}_t\|}=1$. If not perform projection updates, DiZO reverts to standard ZO. \\
\textbf{\uline{Projection clipping.}} To prevent drastic weight changes and maintain training stability, we introduce projection clipping. Specifically, given a clipping range $\tau > 0$, if the projection magnitude $\frac{\bm{\gamma}_t}{\|\Delta \bm{\theta}_t\|} \notin [1-\tau, 1+\tau]$, it is clipped to remain within this interval. This prevents aggressive model adjustments that could destabilize training.

\section{Overhead Analysis}
\label{overheadana}
We simply analyze the computational overhead of our method here and will elaborate further later.


\textbf{Memory overhead.} Our method requires additional memory as it involves storing the pre-trained model and calculating the weight distance gap with the fine-tuned model, which can become costly when scaling to large LLMs. However, in DiZO, we find that projecting only the weight updates of the \emph{Query} and \emph{Value} layers in the attention module, instead of updating all layers, not only reduces memory requirements but also delivers better performance. As a result, we only need to store the weights of these two types of layers from the pre-trained model, accounting for approximately 16.7\% of the parameters in OPT-2.7B, which is a manageable overhead. Similarly, LoRA~\cite{hu2021lora} also focuses on weight decomposition for \emph{Query} and \emph{Value} layers, which echoes our observation.

\textbf{Computational overhead.} 
Our method introduces extra computational cost, as the projection is learned alongside the main optimization (fine-tuning). However, we observe that performing projection learning intermittently, only once every few training iterations, does not compromise performance and significantly reduces the added overhead. This strategy reduces the computational burden while maintaining efficiency, allowing DiZO to achieve throughput comparable to vanilla ZO fine-tuning. 
Additionally, the reduced frequency of projection updates ensures that DiZO remains scalable for larger models and datasets.

\section{Convergence Analysis}
\label{proof1}
In this section, we give a nonconvex convergence guarantee in terms of the expected gradient norm. The bound improves over basic ZO-SGD by replacing the factor $d$ (full dimension) with an effective dimension on the order of $D$ where we denote by $D = \max_{1 \,\le\, \ell \,\le\, L}\,d^{(\ell)}$.

We assume the following, which are standard in ZO analyses:

\begin{assumption}
\label{assumption_smoothness}
$\mathcal{L}$ is $L_f$-smooth, i.e.\ there exists $L_f>0$ such that for all $\bm{\theta}, \bm{\theta}'$,
\[
   \|\nabla \mathcal{L}(\bm{\theta}) \;-\; \nabla \mathcal{L}(\bm{\theta}')\|
   \;\le\; L_f \,\|\bm{\theta} - \bm{\theta}'\|.
\]
\end{assumption}

\begin{assumption}
\label{assumption_projection}
Due to DiZO's layerwise projection step, each $\bm{\theta}^{(\ell)}$ remains in a ball (or line segment) around $\bm{\theta}_0^{(\ell)}$ of radius $R^{(\ell)}$. In particular, define
\[
  \mathcal{S} \;=\; 
  \bigl\{\bm{\theta} \,\mid\, \|\bm{\theta}^{(\ell)} - \bm{\theta}_0^{(\ell)}\|\,\le\, R^{(\ell)} 
          \;\;\forall\,\ell \bigr\}.
\]
The algorithm ensures $\bm{\theta}_t\in \mathcal{S}$ for all iterations $t$.
\end{assumption}

\begin{theorem}
\label{thm:DiZO_convergence}
Under Assumptions~\ref{assumption_smoothness}--\ref{assumption_projection}, suppose DiZO runs for $T$ iterations with step size $\eta = c/\sqrt{T}$ for a sufficiently small constant $c>0$. Then there exist constants such that
\[
  \min_{0 \,\le\, t \,<\, T}
  \mathbb{E}\bigl[\|\nabla \mathcal{L}(\bm{\theta}_t)\|^2\bigr]
  \;\;=\;\;
  O\!\Bigl(\frac{\sqrt{D}}{\sqrt{T}}\Bigr).
\]
\end{theorem}

Note that $D$ can often be much smaller than the total parameter count $\sum_{\ell=1}^L d^{(\ell)}$. This drastically improves the variance bounds in the zeroth-order gradient estimation, leading to a faster convergence rate in practice.
Standard ZO-SGD in $\mathbb{R}^d$ often incurs a factor of $\sqrt{d}$ in its nonconvex stationarity bound, due to estimator variance. 
By restricting each layer $\bm{\theta}^{(\ell)}$ to remain near the pre-trained $\bm{\theta}_0^{(\ell)}$, 
\emph{DiZO} effectively reduces the dimension to $D\!\ll\!d$, improving the rate to $O(\sqrt{D}/\sqrt{T})$.
The full proof and detailed analysis are provided in Appendix~\ref{convergence}.

\section{Experiments}
\label{implementation}

\begin{figure}[t]
\centering
\includegraphics[width=1\linewidth]{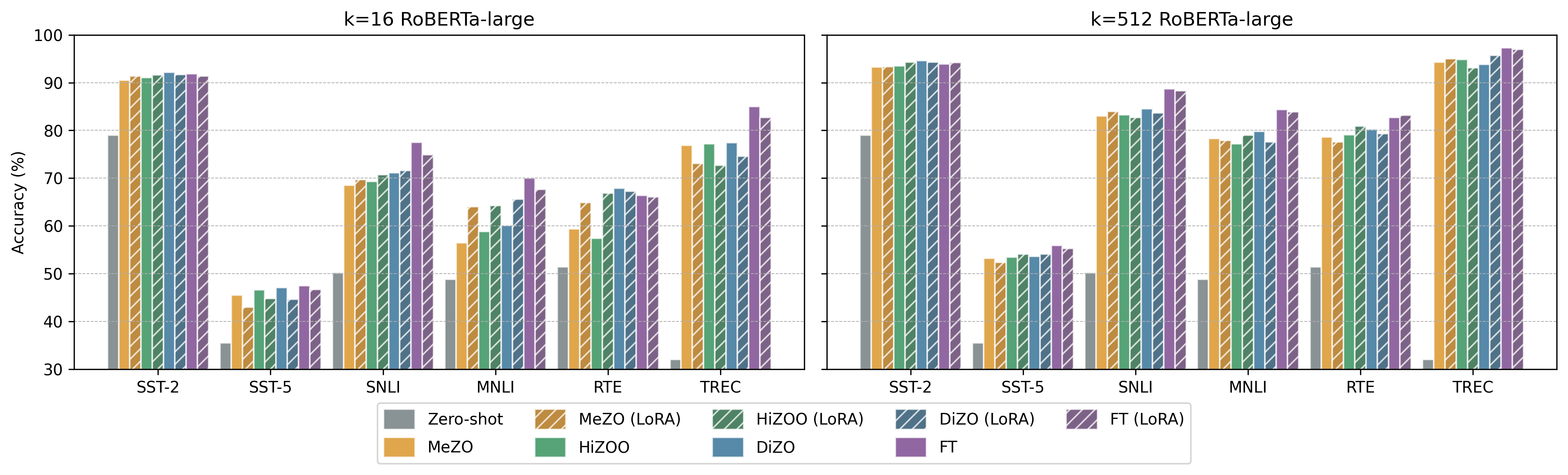}
\caption{Experiments on RoBERTa-large. DiZO outperforms the baselines with and without LoRA. Detailed numbers are presented in Table~\ref{roberta-main}, and the loss trajectory is shown in Figure~\ref{speed_roberta}.
}
\vspace{-10pt}
\label{roberta_figure}
\end{figure}

\subsection{Experimental Settings}

\textbf{Models and datasets.} We evaluate DiZO with various models, including medium-sized masked models~\citep{liu2019roberta} (RoBERTa-large) and large-sized autoregressive models~\citep{zhang2022opt, touvron2023llama} with different size, including OPT-2.7B, OPT-6.7B, OPT-13B, Llama2-7B, Llama3-3B, and Llama3-8B. The total parameter size is ranging from 355M to 13B.  
We evaluate on the SuperGLUE~\citep{wang2019superglue} benchmark, MMLU~\citep{hendrycks2020measuring} and MT-Bench~\citep{zheng2023judging} benchmarks. More details on datasets are shown in Appendix~\ref{dataset}. 

\textbf{Baseline.} We mainly compare with two ZO works, memory-efficient ZO optimization (MeZO)~\citep{malladi2023fine} and Hessian-informed ZO optimization (HiZOO)~\citep{zhao2024second}. MeZO is a fundamental and representative work in ZO LLM fine-tuning but suffers from slow convergence speed. HiZOO
is a recently proposed ZO acceleration for LLM fine-tuning, which leverages the estimated second-order information to speed up. In addition, we also incorporate the parameter-efficient fine-tuning (PEFT) technique LoRA~\citep{hu2021lora}.

\textbf{Evaluation.} 
For SuperGLUE, we follow prior work~\citep{gao2020making, malladi2023fine}, evaluating few-shot and many-shot settings on RoBERTa-large using $k=16$ and $k=512$ samples per class. For each setting, we randomly sample data for training, validation, and testing. For OPT and LLaMA, we use 1000, 500, and 1000 samples respectively. For MMLU and MT-Bench, we adopt the setup in~\citep{pan2024lisa, luo2024badam}, fine-tuning on the Alpaca GPT-4 dataset~\citep{taori2023stanford}. All experiments are conducted on NVIDIA A100 and A6000 GPUs, with results averaged over three trials. Full results are provided in Appendix~\ref{more}.

\subsection{Medium-sized Masked Language Models}

We conduct experiments on RoBERTa-large across three types of datasets and compare DiZO with two ZO baselines. We also explore PEFT by integrating LoRA. Figure~\ref{roberta_figure} presents the results, while Figure~\ref{speed_roberta} shows the trajectory of training loss curves, indicating the convergence speed of DiZO and MeZO. Our key findings are as follows:

\textbf{DiZO greatly increases the convergence speed over MeZO.} By using divergence-driven layer adaptation, the loss curve of DiZO decreases much faster, cutting the required iterations by over 50\% on SST-2, MNLI, and RTE. Moreover, DiZO improves accuracy by 1.7\%, 3.6\%, and 8.5\%.

\textbf{DiZO outperforms MeZO and achieves results on par with full fine-tuning.} From Figure~\ref{roberta_figure}, DiZO consistently surpasses MeZO on all six datasets. Notably, on SST-2 and RTE datasets, DiZO even shows better performance than FO full-parameter fine-tuning, increasing by 0.3\% and 1.5\%.

\textbf{DiZO is effective for both full-parameter fine-tuning and PEFT.} Although DiZO applies projections based on the distance with the pre-trained model, while such prior knowledge does not exist for the decomposed weights of LoRA, it still delivers some gains. 

\subsection{Large Autoregressive Language Models}
\begin{table*}[ht]
\centering
\small
\caption{Experiments results of fine-tuning OPT-2.7B (with 1000 training samples). Better results between MeZO, HiZOO, and DiZO are highlighted in bold.}
\begin{tabular}{lccccccccc}
\toprule
\small
\multirow{2}{*}{\begin{tabular}[c]{@{}l@{}}Dataset\\ Task Type\end{tabular}} & \textbf{SST-2} & \textbf{RTE}  & \textbf{CB}   & \textbf{BoolQ} & \textbf{WSC}  & \textbf{WIC}  & \textbf{MultiRC} & \textbf{SQuAD}       & \textbf{DROP}       \\
                                                                             & \multicolumn{7}{c}{--------------------------classification--------------------------}                             & \multicolumn{2}{c}{------generation------} \\ \midrule
Zero-shot                                                                    & 56.3           & 54.2          & 50.0          & 47.6           & 36.5          & 52.7          & 44.4             & 29.8                 & 10.0                \\
FT                                                                           & 94.2           & 81.2          & 82.1          & 72.2          & 63.8          & 65.8          & 71.6             & 78.4                 & 30.3                \\
LoRA                                                                         & 94.6           & 80.8          & 82.7          & 77.7           & 59.8         & 64.0          & 72.8             & 77.9                 & 31.1                \\ \midrule
MeZO                                                                         & 90.0          & 63.5          & 69.6         & 67.4           & 61.5         & 57.6         & \textbf{58.7}    & 68.7                 & 22.9                \\
HiZOO                                                                        & 90.8          & 60.6         & 70.4         & \textbf{68.0}  & 60.2         & 56.6         & 54.8             & 68.3                 & 23.4                \\
\rowcolor[gray]{.92}DiZO                                & \textbf{92.5}  & \textbf{68.2} & \textbf{71.4} & 67.0           & \textbf{63.4} & \textbf{57.9} & 56.4             & \textbf{69.0}       & \textbf{24.3}       \\ \midrule
MeZO LoRA                                                                    & 91.4          & 66.6          & 71.1          & 67.6          & 59.6         & 57.0          & 57.0             & 70.8                 & 22.5                \\
HiZOO LoRA                                                                   & 90.6          & 65.2         & 71.4         & 67.4          & 52.6         & \textbf{58.8} & \textbf{59.0}    & 71.8                 & 22.7                \\
\rowcolor[gray]{.92}DiZO LoRA                           & \textbf{91.5}  & \textbf{68.4} & \textbf{71.8} & \textbf{70.0}  & \textbf{61.6} & 58.4          & 56.2             & \textbf{74.4}        & \textbf{23.3}       \\
\bottomrule
\end{tabular}

\label{opt2p7b-main}
\end{table*}

\begin{figure*}[htbp]
    \centering
    \vspace{-10pt}
    \begin{minipage}{0.48\textwidth}
        \centering
\captionof{table}{Experiment results on OPT-6.7B (with 1000 training samples).
}
\scalebox{0.7}{
\begin{tabular}{lccccc}
\toprule
\multirow{2}{*}{\begin{tabular}[c]{@{}l@{}}Dataset\\ Task Type\end{tabular}} & \textbf{SST-2} & \textbf{RTE} & \textbf{CB} & \textbf{WSC} & \textbf{SQuAD}   \\
                                                                          & \multicolumn{4}{c}{---------classification---------}                  & --generation-- \\ \midrule
MeZO                                                                      & 90.2     & 73.2  & 71.4  & \textbf{62.2}   & 76.0       \\
HiZOO                                                                     & 90.7     & 74.2   & 71.8 & 62.1   & 77.3       \\
\rowcolor[gray]{.92}DiZO                                                                      & \textbf{91.1}     & \textbf{74.8}   & \textbf{73.2}  & 61.8   & \textbf{78.6}       \\ \midrule
MeZO LoRA                                                                 & 91.6     & 71.2   & 71.4  & 61.8  & 76.3       \\
HiZOO LoRA                                                                & 91.3    & \textbf{71.3}   & 71.4  & 62.1   & 76.1      \\
\rowcolor[gray]{.92}DiZO LoRA                                                                 & \textbf{92.4}     & 70.2   & \textbf{71.8}  & \textbf{62.6}   & \textbf{77.9}       \\
\bottomrule
\end{tabular}
\vspace{-15pt}
}
        
\label{opt6p7b-main}
    \end{minipage}
    \hfill
    \begin{minipage}{0.49\textwidth}
        \centering
        \includegraphics[width=\linewidth]{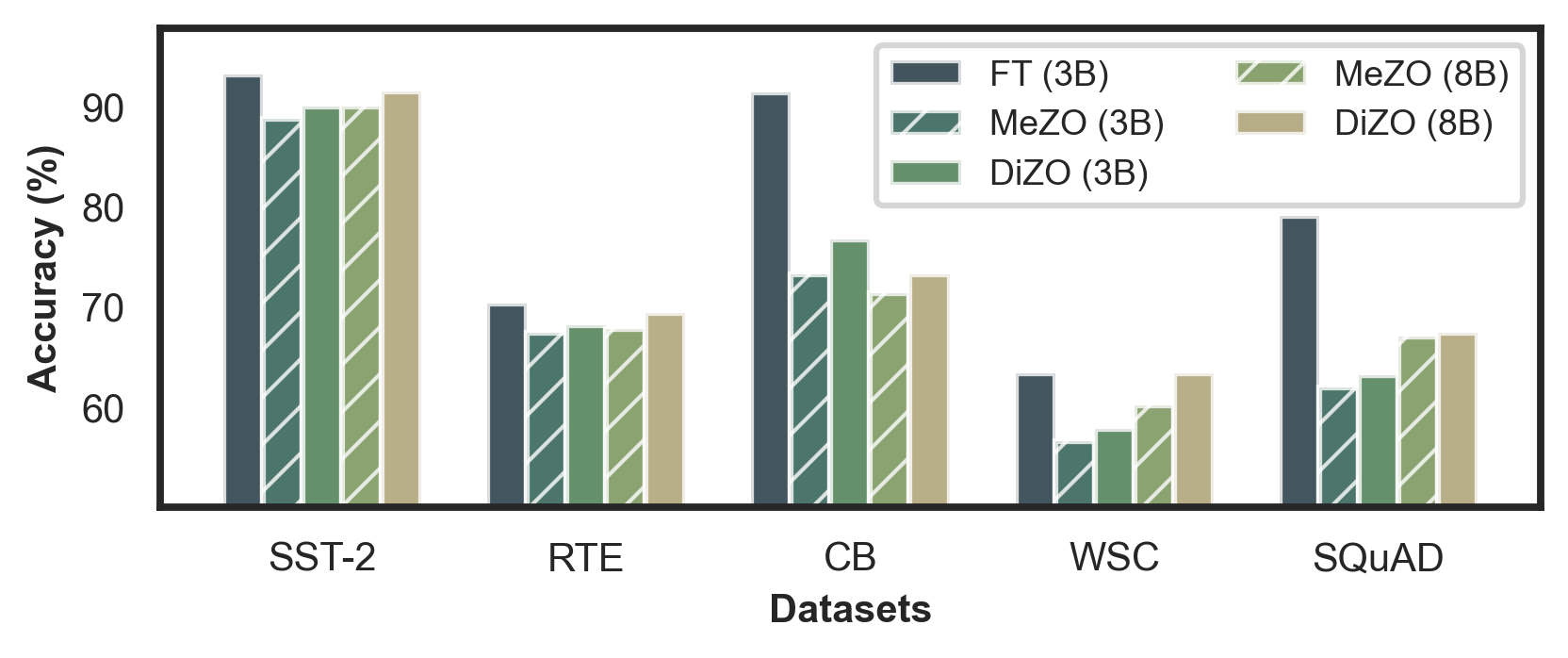}
        \vspace{-16pt}
        \caption{Experiment result on Llama3-3B and Llama3-8B. More results and detailed numbers are shown in Appendix~\ref{Llama}. }
\label{llama_bar}
    \end{minipage}%

\vspace{-10pt}
\end{figure*}

\begin{figure*}[h]
 \centering
 \hspace*{-0.3cm}
 \vspace{-10pt}
 \includegraphics[width=1\linewidth]{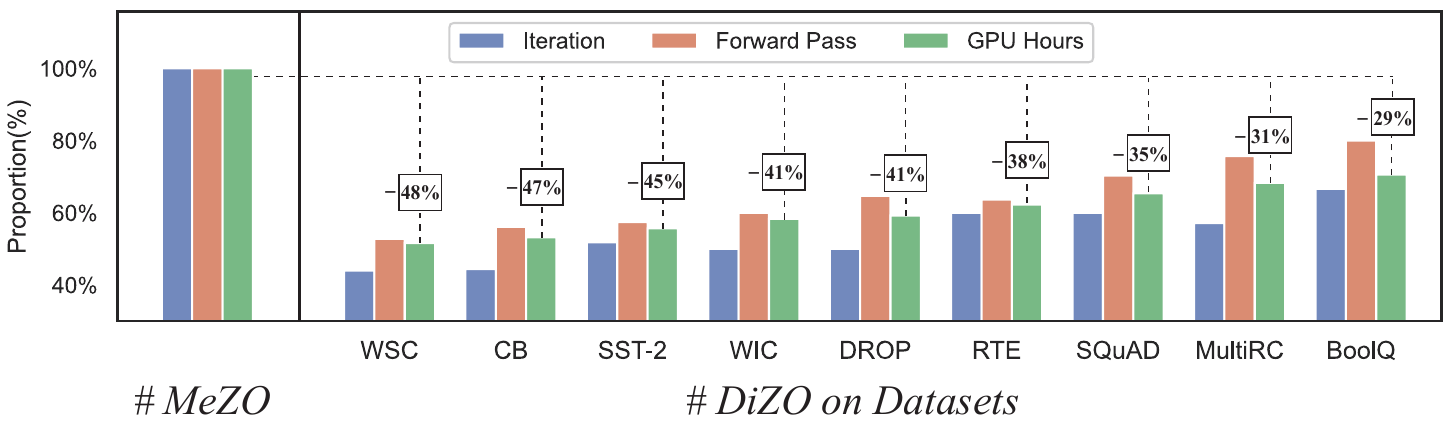}
 \vspace{8pt}
 \caption{
 Comparison between MeZO and DiZO on convergence iteration, forward pass, and training GPU hours across multiple datasets. Results are presented as proportions, with the percentage of saved GPU hours highlighted for each dataset.}
 
 \label{speed_opt}
\end{figure*}

To assess the generalizability of DiZO, we run experiments on the OPT and Llama. The overall results are summarized in Table~\ref{opt2p7b-main}, Table~\ref{opt6p7b-main}, and Figure~\ref{llama_bar} for OPT-2.7B, OPT-6.7B, and Llama series, respectively. We also compare the convergence speeds of DiZO and MeZO on OPT-2.7B across datasets in Figure~\ref{speed_opt}. We highlight key observations from experiments as follows.

\textbf{DiZO significantly reduces training GPU hours over MeZO.} As shown in Table~\ref{speed_opt}, DiZO achieves faster convergence with up to 48\% less GPU time by quickly establishing effective layer-wise divergence. Unlike HiZOO, which reduces iterations but suffers from slow throughput due to costly Hessian estimates, DiZO maintains MeZO-level efficiency with a lightweight projection update using only two forward passes.


\textbf{DiZO outperforms baselines in both standard and parameter-efficient settings.} As shown in Table~\ref{opt2p7b-main}, DiZO consistently outperforms MeZO and HiZOO, with or without LoRA, achieving performance close to FO methods. It ranks first on five of seven classification tasks and leads both text generation tasks. These gains extend to OPT-6.7B (Table~\ref{opt6p7b-main}) and Llama models (Figure~\ref{llama_bar}), highlighting the benefit of layer-wise adaptive updates.


\subsection{Memory and Speed Analysis}
\label{memory_speed}

\begin{table}[htbp]
\small
\centering
\vspace{-15pt}
\caption{Memory utilization and speed test on OPT-2.7B on RTE dataset (180 tokens per example on average). \neutral: partial gradient-free; DiZO\textsuperscript{\textdagger}: learning projection with Adam. For a fair comparison, the speed and memory are measured on the same machine with the same settings.}
\begin{tabular}{lccccccc}
\toprule
Task Type       & \textbf{\begin{tabular}[c]{@{}c@{}}Gradient\\ Free\end{tabular}} & \textbf{\begin{tabular}[c]{@{}c@{}}LoRA\\ Added\end{tabular}} & \textbf{\begin{tabular}[c]{@{}c@{}}Peak \\ Memory\end{tabular}} & \textbf{\begin{tabular}[c]{@{}c@{}}Averaged\\ Memory\end{tabular}} & \textbf{Throughput} & \textbf{\begin{tabular}[c]{@{}c@{}}\#Train \\ Iter.\end{tabular}} & \textbf{\begin{tabular}[c]{@{}c@{}}GPU \\ Hours\end{tabular}} \\ \hline
FT         & \xmark                                            & \xmark                                                  & 62.2 GB                                                      & 62.2 GB                                                        & 1.05 it/s          & 10.0\%                                                            & 16.2\%                                                    \\
LoRA       & \xmark                                            & \cmark                                                  & 42.5 GB                                                      & 42.5 GB                                                        & 2.12 it/s          & 8.3\%                                                        & 6.6\%                                                    \\
DiZO\textsuperscript{\textdagger}       & \neutral                                          & \xmark                                                  & 44.7 GB                                                      & 10.1 GB                                                        & 1.43 it/s          & 33.3\%                                                            & 39.6\%                                                    \\
DiZO LoRA\textsuperscript{\textdagger}  & \neutral                                          & \cmark                                                  & 40.1 GB                                                      & 9.8 GB                                                        & 2.40 it/s          & 26.6\%                                                            & 18.8\%                                                   \\ \midrule
MeZO       & \cmark                                            & \xmark                                                  & 7.8 GB                                                      & 7.8 GB                                                        & 1.70 it/s          & 100.0\%                                                            & 100.0\%                                                    \\
HiZOO      & \cmark                                            & \xmark                                                  & 13.2 GB                                                      & 13.2 GB                                                        & 1.21 it/s          & 63.3\%                                                            & 88.9\%                                                    \\
\rowcolor[gray]{.92}DiZO       & \cmark                                            & \xmark                                                  & 9.5 GB                                                      & 9.5 GB                                                        & 1.54 it/s          & 60.0\%                                                            & 62.3\%                                                    \\ \midrule
MeZO LoRA  & \cmark                                            & \cmark                                                  & 7.7 GB                                                     & 7.7 GB                                                        & 3.10 it/s          & 94.2\%                                                            & 51.6\%                                                    \\
HiZOO LoRA & \cmark                                            & \cmark                                                  & 13.0 GB                                                      & 13.0 GB                                                        & 2.07 it/s          & 80.0\%                                                            & 65.7\%                                                    \\
\rowcolor[gray]{.92}DiZO LoRA  & \cmark                                            & \cmark                                                  & 9.4 GB                                                    & 9.4 GB                                                        & 2.87 it/s          & 66.7\%                                                            & 39.5\%    \\ 
\bottomrule
\end{tabular}
\label{memory_rte}
\end{table}
In this section, we examine the memory utilization and convergence speed of DiZO in comparison with both ZO baselines and FO fine-tuning approaches (with and without LoRA). Table~\ref{memory_rte} presents the results of fine-tuning OPT-2.7B on the RTE dataset, more results are shown in Appendix~\ref{speed}.

From a memory perspective, DiZO avoids backpropagation and memory-heavy activations, cutting memory use by ~90\% compared to FO. Its overhead stems only from storing \emph{Query} and \emph{Value} weights (16.7\% of total). In contrast, HiZOO stores full-layer Hessians, scaling poorly with model size. In terms of convergence speed, DiZO significantly reduces iteration count while maintaining throughput comparable to MeZO, leading to much lower training GPU hours. By comparison, HiZOO achieves less iteration reduction and slows throughput of MeZO by about 1.5× due to Hessian estimation, resulting in only modest savings, or even higher training cost in some cases, such as HiZOO+LoRA on RTE

A notable byproduct of our method is using FO (e.g., with the Adam optimizer) to learn the projections. While this version has memory consumption comparable to LoRA and requires additional training GPU hours, it offers distinct advantages. Since DiZO does not update projections at every iteration, FO-based DiZO exhibits significantly lower average memory usage than FO-based LoRA, with an average memory overhead close to that of the ZO-based DiZO. Although average memory usage may seem less critical in single-process, single-GPU setup, many real-world on-device training scenarios involve multi-process environments~\citep{li2024flexnn, ye2024asteroid}. In such cases, the FO-based DiZO can stagger memory usage phases across processes, enabling parallel operations that purely FO methods cannot achieve. Furthermore, compared with ZO-based DiZO, the FO version reduces extra training GPU hours and delivers better performance. 
These qualities make it particularly appealing for specific on-device training cases.

\subsection{Discussion on Potential Alternatives and Limitations}
\label{discussion}

\textbf{Adaptive learning rate} methods may appear analogous to DiZO at first glance, as both introduce per-layer adaptive control over parameter updates, but their principles differ. Methods like Adam and RMSProp adjust update magnitude based on gradient history, controlling how fast parameters move. DiZO, by contrast, uses geometric constraints to guide parameters toward a learnable target distance from a fixed anchor (the pre-trained model), determining where the parameters move. This projection-based approach enables principled, divergence-aware updates that step-size modulation alone cannot replicate. Additionally, adaptive methods maintain gradient moment estimates, adding memory and computational overhead, particularly in LLMs. We empirically compare DiZO to these methods in Appendix~\ref{alternative}.

\textbf{Line search} could potentially be a simpler method to replace ZO for optimizing projection scalars. However, line search approaches, e.g., backtracking, generally require tuning each layer’s scalar independently, leading to inefficient and unscalable coordinate-wise search. These methods also rely on directional derivatives and assume smooth interactions, which do not generalize well to joint tuning across layers. DiZO avoids these issues by using a ZO-based strategy that updates all scalars simultaneously, stabilized by projection clipping and re-initialization. As shown in Appendix~\ref{alternative}, replacing ZO with line search under the same forward-pass budget leads to worse performance, confirming its inefficiency in this context.

\textbf{Limitations.} While DiZO demonstrates notable improvements in both accuracy and training efficiency, the theoretical foundations behind its design choices remain incomplete. For example, the choice of using a pre-trained model as the anchor point and the selection of which layer to be projected is primarily supported by empirical observations rather than formal justification(detailed ablation study in Appendix~\ref{closer_look}). The absence of a solid theoretical framework to explain why such design yields consistent performance gains leaves open questions about the optimality of the approach. Nevertheless, our findings offer valuable insights and point to promising directions for future research, particularly in developing anchor-guided, adaptive ZO optimization frameworks with stronger theoretical grounding.




\section{Conclusion}

In this paper, we propose a novel layer-wise divergence analysis to reveal the distinct update pattern between FO and ZO methods. Building on these insights, we present DiZO, an enhanced ZO method using divergence-driven layer adaptation to resemble the learning capacity of the FO method. DiZO achieves significant training acceleration and superior performance across diverse tasks and architectures. Moreover, our method can be seamlessly integrated with
PEFT techniques like LoRA for additional speedup. For future work, we plan to explore DiZO in other fields, particularly for fine-tuning large pre-trained vision models.

\newpage

\section{Acknowledgment}

This work was supported by the U.S. National Science Foundation, under Grant No. CCF-2553684, CNS-2312158, No. 1943046.



\bibliography{neurips_2025.bib}
\bibliographystyle{unsrt}

\newpage
\appendix
\onecolumn

\section{Related Work}

\subsection{Memory-efficient Fine-tuning}


Fine-tuning a pre-trained model offers a powerful way to reuse learned representations and reduce training costs compared to building models from scratch, often achieving superior performance~\citep{gururangan2020don,ouyang2022training}. Initially successful in NLP with models like BERT, RoBERTa, and GPT~\citep{devlin2018bert,liu2019roberta,chen2022visualgpt}, fine-tuning has also shown promise in vision tasks such as CLIP and SWAG~\citep{radford2021learning,singh2022revisiting}. 

Despite the success of fine-tuning, its high cost makes it not feasible. Therefore memory efficient fine-tuning method come up. Recent parameter-efficient fine-tuning (PEFT), including LoRA~\citep{hu2021lora}, and prefix tuning~\citep{li2021prefix}, minimize resource needs by updating only a small subset of parameters, preserving most of the pre-trained weights and ensuring valuable knowledge is retained. Low-rank decomposition-based methods, led by LoRA, have achieved remarkable success. The main idea is to minimize resource needs by updating only a small subset of parameters, preserving most of the pre-trained weights and ensuring valuable knowledge is retained. DoRA~\citep{liu2024dora} decomposes the pre-trained weight into two components, magnitude and direction, to enhance both the learning capacity and training stability of LoRA. GaLore~\citep{zhao2024galore} proposed gradient low-rank projection, allows full-parameter learning while retaining the memory advantages of low-rank training. Beside low-rank method, quantization stood out as a promising method to reduce resources utilization. GPT3.int8()~\citep{dettmers2022gpt3} identified the outlier in activation, and include a new mixed-precision decomposition scheme, which isolates the outlier feature dimensions into a 16-bit matrix multiplication while still more than 99.9\% of values are multiplied in 8-bit. Despite training with mix-precision, SmoothQuant~\citep{xiao2023smoothquant} smooths the activation outliers by offline by scaling, migrating the quantization difficulty from activations to weights. Moreover, Outlier Suppression+~\citep{wei2023outlier} and OmniQuant~\citep{shao2023omniquant} apply channel-wise shifting for asymmetry and channel-wise scaling for concentration for int4-level quantization.

\subsection{Zeroth-order Optimization and Acceleration}

ZO optimization emerges as an attractive technique that optimizes the model without backpropagation~\cite{ chen2023deepzero, chen2017zoo, ye2018hessian, verma2023certified, dhurandhar2018explanations, dhurandhar2019model}. Unlike most frequently used FO optimization, which directly obtains and leverages the gradient for optimization, the zeroth-order method utilizes the objective function value oracle only, estimating the gradient by finite differences.
ZO method has a wide range of applications in machine learning fields, including adversarial attack and defense~\cite{chen2017zoo, ye2018hessian, verma2023certified}, machine learning explainability~\cite{dhurandhar2018explanations, dhurandhar2019model}, reinforcement learning~\cite{vemula2019contrasting}, and on-chip training~\cite{gu2021efficient}. 
Recently, the ZO method has been proposed to be leveraged on LLM fine-tuning to address the significant memory usage. \cite{malladi2023fine} proposed MeZO, first scaling ZO optimization to fine-tuning parameter-intensive LLMs, greatly reducing memory utilization. On top of MeZO, \cite{zhao2024second} proposed HiZOO, leveraging the estimated Hessian information for better learning capacity, but reducing the throughput of MeZO to some extent.

ZO optimization, although it significantly saves memory, converges more slowly than FO methods due to higher variance from random search. 
\cite{liu2018zeroth} introduced ZO-SVRG by incorporating variance reduction techniques~\cite{johnson2013accelerating}. \cite{shu2023zeroth} proposed using a Gaussian process to model objective function queries, thereby reducing query complexity and allowing more frequent queries to lower gradient variance. \cite{sener2020learning} performed random search on a learned low-dimensional manifold, reducing the number of needed objective queries. 
However, existing ZO accelerators face two main challenges when adapting to ZO fine-tuning for LLMs. First, these approaches were typically designed for smaller-scale tasks involving fewer parameters and less data, and cannot be directly extended to large-scale LLMs. Second, many prior methods focus on improving query efficiency, whereas recent work has shown that a single query can suffice for LLM fine-tuning~\cite{malladi2023fine}. How to effectively accelerate ZO optimization on large model fine-tuning remains a problem.

Moreover, ZO has several properties that make it well-suited for on-device or edge training scenarios. 1) Memory efficiency: Edge devices such as mobile phones and FPGAs typically offer limited memory resources. ZO significantly reduces memory usage by avoiding activation and gradient storage, making it more deployable in such constrained setting. 2) Forward-only optimization: As ZO only relies on forward passes, it is compatible with existing inference accelerators (e.g., NNAPI on Android, edge TPUs, etc.), which typically lack support for backpropagation. This makes ZO a strong candidate for adapting inference-only hardware for training.


\section{Experiment Settings and Analysis}
\label{imp}
\renewcommand{\thetable}{B.\arabic{table}}
\setcounter{table}{0}

\subsection{Datasets and Evaluation}
\label{dataset}

\begin{table*}[htbp]
\small
\centering
\caption{The hyperparameter for experiments. For DiZO and DiZO LoRA, we only show the setting of extra hyperparameters, and have the same setting in other common hyperparameters with MeZO and MeZO LoRA respectively. 
}
\vspace{5pt}
\begin{tabular}{lrc}
\hline
Experiment            & Hyperparameters                     & Values                                                                                   \\ \hline
\multirow{3}{*}{FT}   & Batch size                          & 8                                                                                        \\
                      & Learning rate                       & \{1e-5, 5e-5\}                                                                           \\
                      & Lr schedule                         & \begin{tabular}[c]{@{}c@{}}Constant for RoBERTa\\ Linear for OPT and Llama\end{tabular}  \\ \hline
\multirow{4}{*}{MeZO} & Batch size                          & \{64, 16\}                                                                               \\
                      & Learning rate $\eta$ (Lr)         & \{1e-6, 5e-7\}                                                                           \\
                      & $\epsilon$                          & 1e-3                                                                          \\
                      & Lr schedule                         & \begin{tabular}[c]{@{}c@{}}Constant for RoBERTa \\ Linear for OPT and Llama\end{tabular} \\ \hline
\multirow{4}{*}{MeZO LoRA} & Batch size                          & \{64, 16\}                                                                               \\
                      & Learning rate $\eta$ (Lr)         & \{1e-4, 5e-5\}                                                                           \\
                      & $\epsilon$                          & 1e-2                                                                           \\
                      & Lr schedule                         & \begin{tabular}[c]{@{}c@{}}Constant for RoBERTa \\ Linear for OPT and Llama\end{tabular} \\ \hline
\multirow{4}{*}{DiZO (LoRA)} & Projection update cycle & \{50, 100, 200, 400\}                                                                    \\
                      & Smoothing scalar $\epsilon^{'}$         & \{1e-1, 5e-2\}                                                                            \\

                      & Clip range $\tau$                          & \{0.1, 0.2, 0.3\}                                                                     \\ \hline
\end{tabular}

\label{setting}
\end{table*}
For the RoBERTa-large model, we use the following classification datasets: SST-2~\cite{socher2013recursive}, SST-5~\cite{socher2013recursive}, SNLI~\cite{bowman2015large}, TREC~\cite{voorhees2000building}, MNLI~\cite{yao2020pyhessian}, and RTE~\cite{dagan2005pascal,bar2006second,bentivogli2009fifth,giampiccolo2007third}. Following previous studies, we cap the test set size at 1000 samples. Two training settings are considered: $k=16$ and $k=512$, where we randomly select 16 or 512 samples per class for both training and validation.

For the OPT and Llama series models, we use the SuperGLUE benchmark~\cite{wang2019superglue}, which includes RTE~\cite{dagan2005pascal,bar2006second,bentivogli2009fifth,giampiccolo2007third}, CB~\cite{de2019commitmentbank}, BoolQ~\cite{clark2019boolq}, WIC~\cite{pilehvar2018wic}, WSC~\cite{levesque2012winograd}, and MultiRC~\cite{khashabi2018looking}. We also include SST-2~\cite{socher2013recursive} and two question answering datasets, SQuAD~\cite{rajpurkar2016squad} and DROP~\cite{dua2019drop}. For each of these datasets, we randomly sample 1000 instances for training, 500 for validation, and 1000 for testing.

\subsection{Hyperparameter Setting}

We use the hyperparameters in Table~\ref{setting} for experiments on RoBERTa-large, OPT-series, and Llama-series models. Specifically, the choice of clip range did not significantly impact the performance. The selection of the projection update cycle and scalar for projection affects the performance somewhat. Generally, for datasets that need larger iterations for convergence, or for these harder datasets, DiZO prefers a larger update cycle, while for those less complicated datasets, DiZO benefits from a smaller update cycle.

\section{Ablation study on DiZO}
\label{closer_look}

\renewcommand{\thetable}{C.\arabic{table}} 
\renewcommand{\thefigure}{C.\arabic{figure}} 
\setcounter{table}{0}  
\setcounter{figure}{0}  

\subsection{Ablation for Projection Layers Selection}
\label{ablation_l}

Instead of applying projections to all layers, which would require storing the entire pre-trained model, we focus only on projecting the weights of the \emph{Query} and \emph{Value} in the attention modules. As shown in Table~\ref{ablation_layer}, this strategy achieves the best trade-off between the overall performance and extra storage requirements, does not reduce the performance and only 16.7\% of the parameters of the pre-trained model are needed to store. A Similar strategy has also been adopted in LoRA~\cite{hu2021lora}.

\begin{table*}
\small
    \centering
    \caption{Ablation study for selecting which layers to project. The highlighted line with a blue rectangle is the setting used in DiZO. Extra memory indicates the extra memory needed due to pre-trained model storing. Attn\_Q: attention Query layer; Attn\_V: attention Value layer; Attn\_K: attention Key layer; Attn\_O: attention output projection; Dense: dense fully connected layer.}
    \vspace{5pt}
    \begin{tikzpicture}
        \node (table) [inner sep=0pt] { 
            \begin{tabular}{cccccc|ccc}
                \toprule
                \textbf{Attn\_Q} & \textbf{Attn\_V} & \textbf{Attn\_K} & \textbf{Attn\_O} & \textbf{Dense} & \textbf{Extra memory} & \textbf{SST-2} & \textbf{RTE} & \textbf{SQuAD} \\ \hline
                \cmark  & \cmark  & \cmark  & \cmark  & \cmark  & 100\%       & 91.7  & 68.4 & 67.3  \\
                \cmark  & \cmark  & \cmark  & \cmark  & \xmark  & 33.3\%      & 92.2  & 67.9 & 69.2  \\
                \cmark  & \cmark  & \cmark  & \xmark  & \xmark  & 25.0\%      & 91.9  & 67.1 & 68.1  \\
                \cmark  & \cmark  & \xmark  & \xmark  & \xmark  & 16.7\%      & 92.5  & 68.2 & 69.0  \\
                \cmark  & \xmark  & \xmark  & \xmark  & \xmark  & 8.4\%      & 90.5  & 64.9 & 66.5  \\
                \bottomrule
            \end{tabular}
        };

\draw[blue!60!white, very thick] (-6.6, -0.70) rectangle (6.6,-0.4);
    \end{tikzpicture}
    
    \label{ablation_layer}
\end{table*}

\subsection{Ablation for Strategies in ZO Projection Learning}
\label{ablation}

\begin{figure}[h]
\centering
\begin{minipage}{0.48\textwidth}
\small
\setlength{\tabcolsep}{3pt}
\begin{tabular}{lcc|ccc}
\toprule
\textbf{Type}                 & \textbf{Re-init} & \textbf{Clipping} &\textbf{SST-2} & \textbf{SNLI}  & \textbf{TREC} \\ \hline
MeZO                  &-         &-          & 90.5  & 66.0 & 76.9  \rule{0pt}{2.5ex} \\ \midrule
\multirow{3}{*}{DiZO} &\xmark         &\cmark          & 88.6  & 64.2 & 73.8  \\
                      &\cmark         &\xmark          & 90.9  & 56.2 & 61.2  \\
                      &\cmark         &\cmark          & 92.2  & 71.6 & 77.4  \\
                      \bottomrule
\end{tabular}
\vspace{15pt}

\end{minipage}
\begin{minipage}{0.49\textwidth}
        \centering
        \scalebox{0.98}{
        \includegraphics[width=\linewidth]{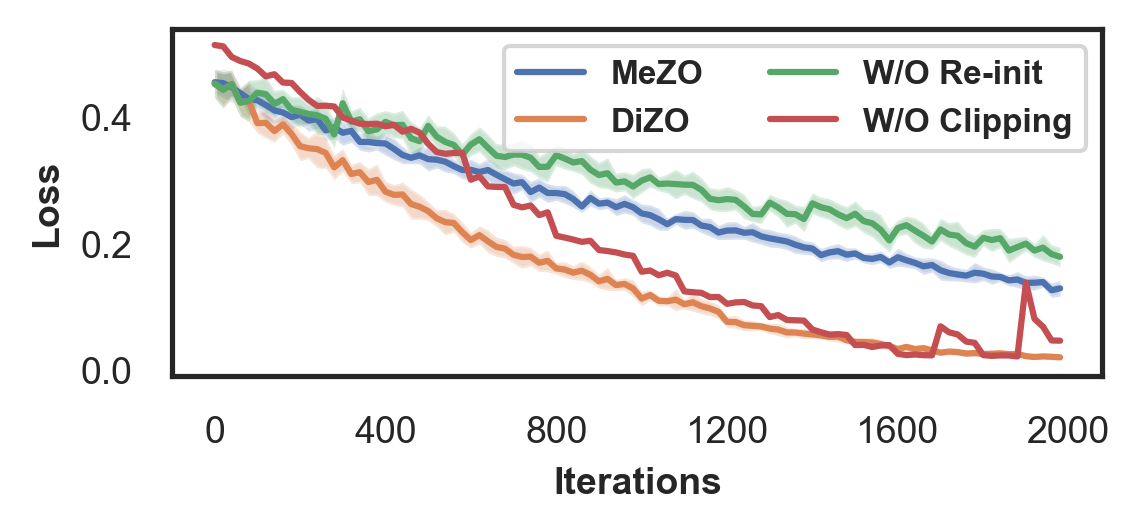}
        }
        \vspace{-14pt}
    \end{minipage}%
    \vspace{5pt}
    \caption{Ablation study for the two strategies: re-initialization and projection clipping, which is conducted on RoBERTa-large ($k=16$). Left: overall results when ablating the strategies. Right: loss curve when ablating the strategies.}
\label{ablation_strategy}
\end{figure}

As discussed in Section~\ref{ZO}, we introduce two strategies, \emph{re-initialization} (Re-init) and \emph{projection clipping} (Clipping), to enhance projection learning and improve the stability of fine-tuning. The ablation results for these strategies, along with the corresponding loss curves, are shown in Figure~\ref{ablation_strategy}.

Overall (left in Figure~\ref{ablation_strategy}), omitting either Re-init or Clipping significantly diminishes the benefits of DiZO, with MeZO outperforming DiZO in these cases. Specifically, without Re-init, accuracy drops sharply, falling below MeZO. Similarly, without Clipping, while DiZO slightly outperforms MeZO on simpler datasets like SST-2, it suffers from severe model collapse on more challenging datasets, leading to a significant decline in accuracy.

From the loss curve trajectory (right in Figure~\ref{ablation_strategy}), without Re-init, DiZO loses its advantage in training acceleration, as the loss curve becomes noticeably slower to decrease. Without Clipping, the loss curve exhibits significant oscillations during certain training steps. This instability arises when projections are optimized to unsuitable values, such as extremely large or small magnitudes. These inappropriate projections cause substantial changes in model weights, leading to pronounced oscillations in the loss.

\subsection{Ablation for the Selection of the Anchor}
\label{anchor_selection}

We conduct an experiment to illustrate the effect of other anchor points beyond the pre-trained weights, the results are shown in Table~\ref{ablation_projection}. In conclusion, using $\bm{0}$ or $\bm{\theta_{t-1}}$ significantly reducing the benefit of our method. Specifically, using $\bm{0}$ as an anchor yields similar results to using $\theta_{t-1}$ in terms of GPU hours, but causes unstable training, the accuracy decreases at the later training stage, and causes the results to be even worse than MeZO. Therefore, the effectiveness of our method is inseparable from the choice of anchors. Selecting a more robust anchor could not only improve accuracy but also the convergence speed.

\begin{table}
\centering
\caption{Comparison on conducting projection on learning rate (LR) or use weight at $(t-1)$-th iteration $\bm{\theta}_{t-1}$ instead of the weight of the pre-trained model $\bm{\theta_{0}}$ as the base of projection. Results are obtained by fine-tuning OPT-2.7B.}
\vspace{5pt}
\begin{tabular}{lcccccc}
\toprule
\multirow{2}{*}{Anchor} & \multicolumn{2}{c}{\textbf{SST-2}}                         & \multicolumn{2}{c}{\textbf{RTE}}                           & \multicolumn{2}{c}{\textbf{SQuAD}}                         \\ \cline{2-7} 
                                                                             & Acc. & \begin{tabular}[c]{@{}c@{}}GPU\\ Hours\end{tabular} & Acc. & \begin{tabular}[c]{@{}c@{}}GPU\\ Hours\end{tabular} & F1.  & \begin{tabular}[c]{@{}c@{}}GPU\\ Hours\end{tabular} \\ \hline
NA (MeZO)                                                                         & 90.0 & 100.0\%                                                & 63.5 & 100.0\%                                                & 68.7 & 100.0\%                                                \\
$\bm{0}$                                                                & 86.9 & 85.7\%                                                & 58.4 & 91.0\%                                                & 62.2 & 85.8\%                                                \\
$\bm{\theta}_{t-1}$ projection                                                             & 90.7 & 87.8\%                                                & 64.5 & 90.3\%                                                & 67.2 & 88.4\%                                                \\
\rowcolor[gray]{.92}DiZO                                                                         & 92.5 & 55.7\%                                                & 68.2 & 62.3\%                                                & 69.0 & 65.4\%                                                \\ \bottomrule
\end{tabular}
\label{ablation_projection}
\end{table}

\subsection{Ablation on Hyperparameter Setting}
\begin{figure}[htbp]
  \centering

  \begin{minipage}{0.32\linewidth}
    \centering
    \includegraphics[width=\linewidth]{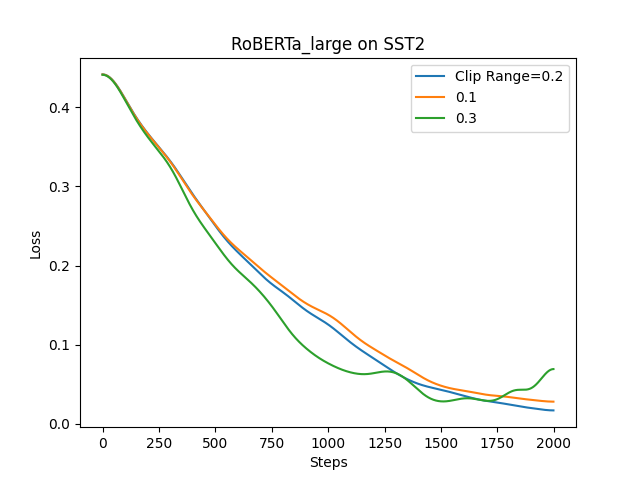}
  \end{minipage}
  \begin{minipage}{0.32\linewidth}
    \centering
    \includegraphics[width=\linewidth]{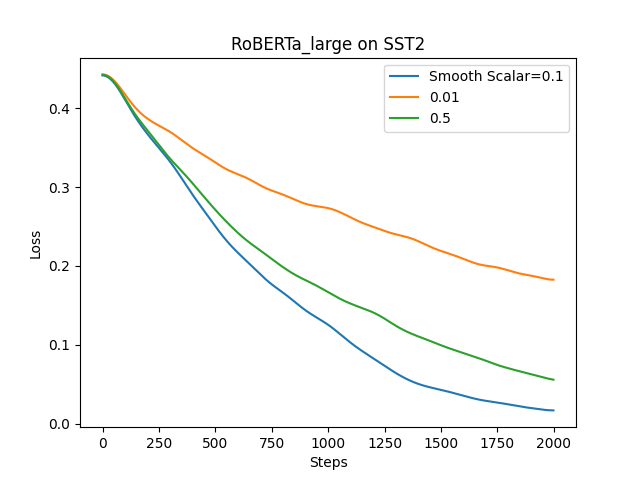}
  \end{minipage}
  \begin{minipage}{0.32\linewidth}
    \centering
    \includegraphics[width=\linewidth]{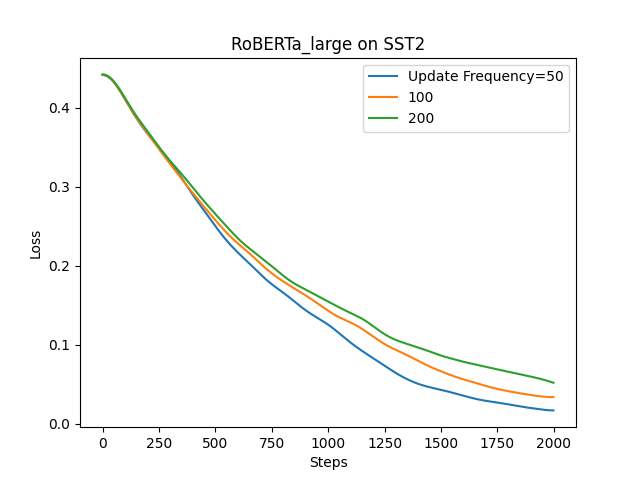}
  \end{minipage}

  \qquad

  \begin{minipage}{0.32\linewidth}
    \centering
    \includegraphics[width=\linewidth]{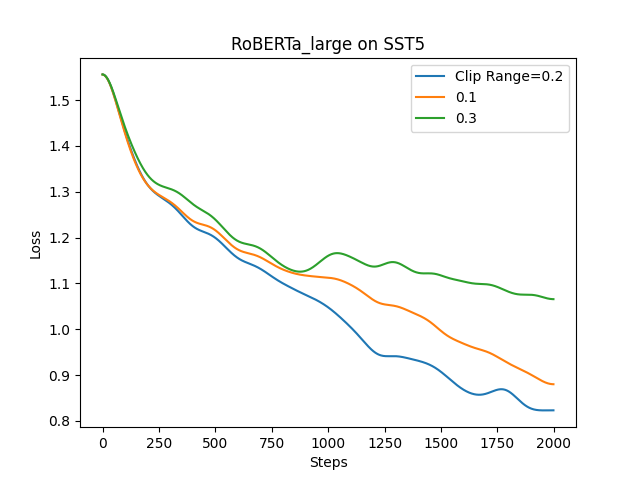}
  \end{minipage}
  \hfill
  \begin{minipage}{0.32\linewidth}
    \centering
    \includegraphics[width=\linewidth]{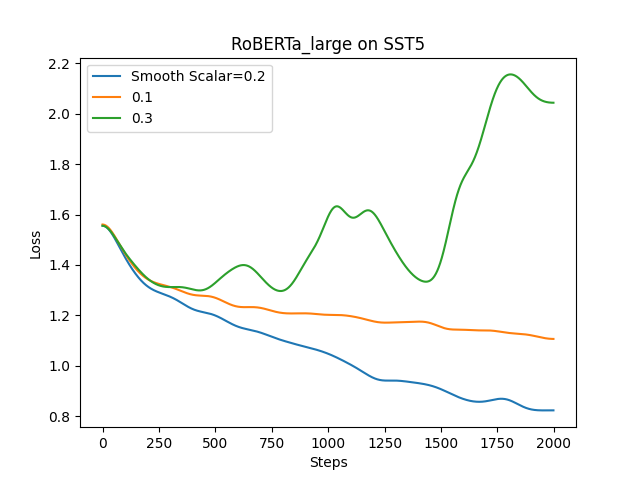}
  \end{minipage}
  \hfill
  \begin{minipage}{0.32\linewidth}
    \centering
    \includegraphics[width=\linewidth]{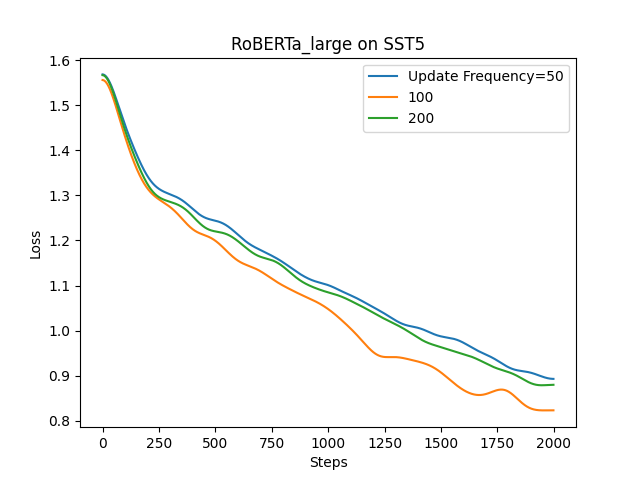}
  \end{minipage}

  \caption{Hyperparameter sensitive testing, including clip range, smooth scalar, and update frequency. Evaluated by training Roberta-large on SST2 and SST5 datasets.}
  \label{hyper_ablation}
\end{figure}

We conduct experiments to investigate the effect of different hyperparameter settings, including different clip ranges, smoothing scalars, and projection update frequencies. Results are obtained by fine-tuning RoBERTa-large on SST-2 and SST-5 with 3 different settings for the 3 hyperparameters, as shown in Figure~\ref{hyper_ablation}. In conclusion, for easier classification tasks like SST-2, the weight of the model changes fast, and so as the convergence speed, therefore, we can apply more aggressive projection strategies, apply a larger clip range, a larger smoothing scalar, and update the projection more frequently. In contrast, for the rather more difficult tasks, a more conservative projection strategy is preferred.

\subsection{Quantization of Anchor (Pre-trained) Model}

Our method introduce a anchor model (i.e., the pre-trained model) for projection, roughly 16\% of the pretrained parameters must still reside in memory. To further reduce extra memory cost, we explored anchor compression via quantization of the pretrained Query and Value matrices to 8-bit and 4-bit precision~\citep{shao2023omniquant}. As shown in Table~\ref{quant_anchor}, DiZO can effectively incorporate with the quantization technique, still preserving advantages in both accuracy and GPU hours. Exploring more advanced quantization or compression methods to further improve anchor efficiency while maintaining performance is an important avenue for future work.

\begin{table}[htbp]
\centering
\caption{DiZO with quantized anchor.}
\begin{tabular}{lcc|cc}
\toprule
\textbf{Method} & \multicolumn{2}{c|}{\textbf{SST-2}} & \multicolumn{2}{c}{\textbf{RTE}} \\
 & \textbf{Acc} & \textbf{GPU hours} & \textbf{Acc} & \textbf{GPU hours} \\
\hline
MeZO & 90.0 & 100\% & 63.5 & 100\% \\
DiZO (8-bits) & 92.2 & 63\% & 67.2 & 71\% \\
DiZO (4-bits) & 91.7 & 67\% & 65.2 & 68\% \\
DiZO & 92.5 & 56\% & 68.4 & 62\% \\
\bottomrule
\end{tabular}
\label{quant_anchor}
\end{table}

\section{Do Other Alternative Strategies Work?}
\label{alternative}

\renewcommand{\thetable}{D.\arabic{table}} 
\renewcommand{\thefigure}{D.\arabic{figure}} 
\setcounter{table}{0}  
\setcounter{figure}{0}  

As discussed in Section~\ref{discussion}, we compare with two representative straightforward alternative strategies, Adam and Backtracking linesearch, to highlight the effectiveness of our method.

\textbf{Adam} as a representative of learnable learning rate methods, Adam leverages first- and second-moment estimates of historical gradients to adaptively modulate the update magnitude. However, incorporating Adam into ZO optimization poses significant practical challenges. Storing gradient moments leads to over 3$\times$ memory overhead compared to ZO-SGD, as shown in Table~\ref{adam_results}. MeZO~\citep{malladi2023fine} attempts to address this by recomputing moment statistics on the fly rather than storing them. However, this strategy is also impractical as the total computational overhead increases quadratically with the training iterations, e.g., 4.3$\times$ more FLOPs for only 2K iterations. Given that zeroth-order methods often require tens of thousands of iterations to converge, both strategies render Adam either memory-intensive or computationally impractical at scale. In contrast, DiZO with SGD achieves higher accuracy with significantly lower FLOPs and comparable memory usage.

\begin{table}
\centering
\caption{Results on fine-tuning OPT-2.7B on SST2 when using Adam as the optimizer. Adam is either memory-intensive or introduces a lot of extra computational overhead.}
\begin{tabular}{lccccc}
\toprule
\textbf{Method} & \textbf{Optimizer} & \textbf{Acc.} & \textbf{Training FLOPs} & \textbf{Memory} & \textbf{Iter.} \\ \midrule
MeZO            & SGD                & 85.2          & 100\%                   & 6.8 GB          & 2K             \\
MeZO            & Adam (Recompute)   & 86.1          & 431\%                   & 6.8 GB          & 2K             \\
MeZO            & Adam (Store)       & 86.1          & 100.2\%                 & 17.6GB          & 2K             \\ \midrule
DiZO            & \textbf{SGD}       & \textbf{86.3} & \textbf{61\%}           & 7.5GB           & 1K             \\
DiZO            & SGD                & 89.8          & 122\%                   & 7.5GB           & 2K            \\
\bottomrule
\end{tabular}

\label{adam_results}
\end{table}

\textbf{Line Search} appears to be a simpler alternative for tuning the projection scalar. However, it requires freezing all other layers when optimizing the scalar for a particular layer, resulting in significant computational overhead. To empirically validate this limitation, we replace the ZO-based projection search in DiZO with Armijo-style backtracking line search. For fairness, the line search is initialized at $(1+\tau)\times\text{original learning rate}$, matching the scale used in DiZO. The results in Table~\ref{linesearch} show that, under the same forward pass budget, backtracking performs notably worse than the original MeZO. Furthermore, even when increasing the number of forward passes to 7800, it only matches MeZO's performance and still falls short of DiZO. These results confirm that traditional line search is not only less effective but also less efficient in the context of high-dimensional, layer-wise projection tuning.

\begin{table}
\centering
\caption{Results on fine-tuning OPT-2.7B on SST2. For fair comparison, we use Armijo-style backtracking line search to replace ZO in our method.}
\begin{tabular}{lcc}
\toprule
\textbf{Searching Strategy} & Forward Pass & Accuracy \\ \midrule
NA (MeZO)                   & 4000         & 85.2     \\
Backtracking                & 4000         & 81.6     \\
Backtracking                & 7800         & 85.1     \\
ZO (DiZO)                   & 4000         & 89.3     \\
\bottomrule
\end{tabular}

\label{linesearch}
\end{table}

\section{More Experimental Results}
\label{more}

\renewcommand{\thetable}{E.\arabic{table}} 
\renewcommand{\thefigure}{E.\arabic{figure}} 
\setcounter{table}{0}  
\setcounter{figure}{0}  

In this section, we provide a comprehensive presentation of our results across various datasets and models to complement the main paper. Specifically, the results include:

\begin{itemize}
    \item Detailed accuracy number and trajectory of training loss on RoBERTa-large (Table~\ref{roberta-main} and Figure~\ref{speed_roberta}).
    \item More memory and speed results of fine-tuning OPT-2.7B on SST-2 and SQuAD datasets (Table~\ref{memory_speed_sst2} and Table~\ref{memory_squad}).
    \item Results on larger model, OPT-13B (Table~\ref{OPT13B}).
    \item Results on Llama3-3B and Llama-8B (Table~\ref{Llama-3B} and Table~\ref{Llama-8B}).
    \item Results on more challenging benchmark MMLU and MT-Bench (Table~\ref{mmlu_llama2} and Table~\ref{mmlu_llama3}).
\end{itemize}

\subsection{RoBERTa-large Experiments}

\begin{figure}
 \centering
 \includegraphics[width=\linewidth]{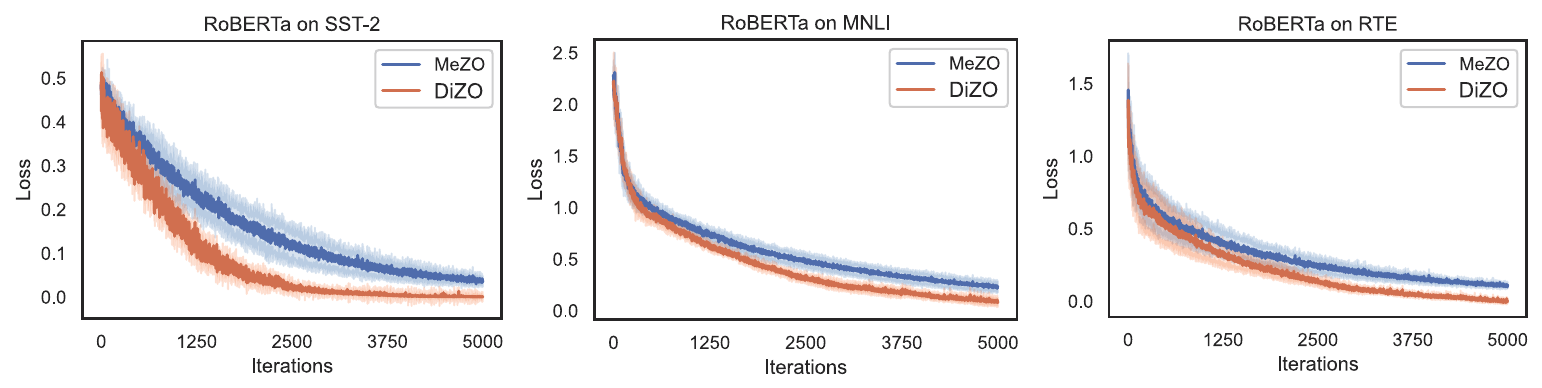}
 \caption{Trajectory of training loss curves when using MeZO and DiZO to fine-tune Roberta-large on SST-2, MNLI, and RTE.}
 \label{speed_roberta}
\end{figure}

Table~\ref{roberta-main} reports the corresponding detailed numbers from Figure~\ref{roberta_figure}, and Figure~\ref{speed_roberta} shows the trajectory of training loss.
\begin{table*}[htbp]
\centering
\small
\setlength{\tabcolsep}{10pt}
\caption{Experiment results on RoBERTa-large (350M) on six classification datasets. Results of the baseline methods MeZO and MeZO LoRA are taken from~\cite{malladi2023fine}. All reported numbers are averaged accuracy with standard deviation shown. Better results between MeZO and DiZO are highlighted in bold.}
\vspace{5pt}
\scalebox{0.9}{
\begin{tabular}{lcccccc}
\toprule
\multirow{2}{*}{\begin{tabular}[c]{@{}l@{}}Dataset\\ Task Type\end{tabular}} & \textbf{SST-2}       & \textbf{SST-5}       & \textbf{SNLI} & \textbf{MNLI} & \textbf{RTE} & \textbf{TREC} \\
                                                                          & \multicolumn{2}{c}{-------sentiment-------} & \multicolumn{3}{c}{----------language inference----------} & --topic--     \\ \hline
Zero-shot                                                                 & 79.0                 & 35.5                 & 50.2          & 48.8          & 51.4         & 32.0    \rule{0pt}{2.5ex}      \\ \hline
\multicolumn{7}{c}{Gradient-free methods: $k=16$}  \rule{0pt}{2.5ex}                                                                                                                                      \\ \hline
MeZO                                                                      & 90.5 (1.2)           & 45.5 (2.0)           & 68.5 (3.9)    & 56.5 (2.5)    & 59.4 (5.3)   & 76.9 (2.7)    \\
MeZO LoRA                                                                 & 91.4 (0.9)           & 43.0 (1.6)           & 69.7 (6.0)    & 64.0 (2.5)    & 64.9 (3.6)   & 73.1 (6.5)    \\
HiZOO                                                                      & 91.1 (1.6)           & 46.1 (1.3)           & 69.3 (3.1)    & 58.8 (3.1)    & 57.4 (6.2)   & 77.2 (2.2)    \\
HiZOO LoRA                                                                 & 91.4 (0.9)           & 44.8 (1.5)           & 70.7 (5.2)    & 64.3 (2.8)    & 66.9 (3.2)   & 72.7 (7.3)    \\
\rowcolor[gray]{.92}DiZO                                                                      & \textbf{92.2} (0.9)           & \textbf{47.1} (1.3)           & 71.0 (3.1)    & 60.1 (3.5)    & \textbf{67.9} (4.7)   & \textbf{77.4} (2.4)    \\
\rowcolor[gray]{.92}DiZO LoRA                                                                 & 91.7 (0.8)           & 44.6 (1.7)           & \textbf{71.6} (3.8)    & \textbf{65.6} (2.8)    & 67.3 (3.9)   & 74.6 (4.3)    \\ \hline
\multicolumn{7}{c}{Gradient-based methods: $k=16$} \rule{0pt}{2.5ex}                                                                                                                                      \\ \hline
FT                                                                        & 91.9 (1.8)           & 47.5 (1.9)           & 77.5 (2.6)    & 70.0 (2.3)    & 66.4 (7.2)   & 85.0 (2.5)    \\
FT LoRA                                                                   & 91.4 (1.7)           & 46.7 (1.1)           & 74.9 (4.3)    & 67.7 (1.4)    & 66.1 (3.5)   & 82.7 (4.1)    \\ \hline \hline
\multicolumn{7}{c}{Gradient-free methods: $k=512$} \rule{0pt}{2.5ex}                                                                                                                                       \\ \hline
MeZO                                                                      & 93.3 (0.7)           & 53.2 (1.4)           & 83.0 (1.0)    & 78.3 (0.5)    & 78.6 (2.0)   & 94.3 (1.3)    \\
MeZO LoRA                                                                 & 93.4 (0.4)           & 52.4 (0.8)           & 84.0 (0.8)    & 77.9 (0.6)    & 77.6 (1.3)   & 95.0 (0.7)    \\
HiZOO                                                                      & 93.5 (0.4)           & 53.5 (1.2)           & 83.3 (1.4)    & 77.2 (1.5)    & 79.1 (1.2)   & 94.9 (1.7)    \\
HiZOO LoRA                                                                 & 94.3 (0.5)           & 54.1 (0.6)           & 82.7 (1.8)    & 79.0 (0.8)    & \textbf{80.9} (1.6)   & 93.1 (0.5)    \\
\rowcolor[gray]{.92}DiZO                                                                      & \textbf{94.6} (0.1)           & 53.6 (1.7)           & \textbf{84.5} (0.6)    & \textbf{79.8} (0.9)    & 80.3 (1.8)   & 93.8 (1.5)    \\
\rowcolor[gray]{.92}DiZO LoRA                                                                 & 94.3 (0.3)           & \textbf{54.1} (1.4)           & 83.7 (1.1)    & 77.6 (0.4)    & 79.3 (1.4)   & \textbf{95.7} (0.9)    \\ \hline
\multicolumn{7}{c}{Gradient-based methods: $k=512$} \rule{0pt}{2.5ex}                                                                                                                                     \\ \hline
FT                                                                        & 93.9 (0.7)           & 55.9 (0.9)           & 88.7 (0.8)    & 84.4 (0.8)    & 82.7 (1.4)   & 97.3 (0.2)    \\
FT LoRA                                                                   & 94.2 (0.2)           & 55.3 (0.7)           & 88.3 (0.5)    & 83.9 (0.6)    & 83.2 (1.3)   & 97.0 (0.3)   \\
\bottomrule
\end{tabular}
}

\label{roberta-main}
\end{table*}

\subsection{More Memory and Speed Analysis}
\label{speed}
We present the memory and speed results for OPT-2.7B on the SST-2 and SQuAD datasets in Table~\ref{memory_speed_sst2} and Table~\ref{memory_squad}, respectively. DiZO significantly reduces the number of required iterations while maintaining throughput comparable to MeZO, leading to substantially fewer training GPU hours. In contrast, HiZOO achieves only modest iteration savings and further reduces the throughput of MeZO by approximately 1.5× due to its reliance on second-order information estimation. As a result, HiZOO offers only a slight improvement over MeZO in terms of training GPU hours. In some cases, such as HiZOO combined with LoRA on SQuAD, it even consumes more training GPU hours than MeZO with LoRA.

\begin{table}
\centering
\small
\caption{Memory utilization and speed test on OPT-2.7B on SST-2 dataset (35 tokens per example on average). \neutral: partial gradient-free; \cmark: gradient-free; \xmark: gradient-based. DiZO\textsuperscript{\textdagger}: searching projection with Adam.}
\vspace{5pt}
\scalebox{1}{
\begin{tabular}{lccccccc}
\toprule
Task Type       & \textbf{\begin{tabular}[c]{@{}c@{}}Gradient\\ Free\end{tabular}} & \textbf{\begin{tabular}[c]{@{}c@{}}LoRA\\ Added\end{tabular}} & \textbf{\begin{tabular}[c]{@{}c@{}}Peak \\ Memory\end{tabular}} & \textbf{\begin{tabular}[c]{@{}c@{}}Averaged\\ Memory\end{tabular}} & \textbf{Throughput} & \textbf{\begin{tabular}[c]{@{}c@{}}\#Train \\ Iter.\end{tabular}} & \textbf{\begin{tabular}[c]{@{}c@{}}GPU \\ Hours\end{tabular}} \\ \hline
FT         & \xmark                                            & \xmark                                                  & 45.4 GB                                                     & 45.4 GB                                                        & 1.81 it/s          & 9.3\%                                                            & 16.8\%                                                    \\
LoRA       & \xmark                                            & \cmark                                                  & 18.4 GB                                                     & 18.4 GB                                                        & 4.50 it/s          & 5.6\%                                                            & 4.3\%                                                    \\
DiZO\textsuperscript{\textdagger} (w. FO)       & \neutral                                          & \xmark                                                  & 17.8  GB                                                     & 15.7 GB                                                        & 2.63 it/s          & 33.3\%                                                            & 41.5\%                                                    \\
DiZO LoRA\textsuperscript{\textdagger}  & \neutral                                          & \cmark                                                  & 16.1 GB                                                     & 14.7 GB                                                        & 4.16 it/s          & 22.2\%                                                            & 17.5\%                                                    \\ \hline
MeZO       & \cmark                                            & \xmark                                                  & 6.8 GB                                                      & 6.8 GB                                                        & 3.28 it/s          & 100.0\%                                                            & 100.0\%                                                    \\
HiZOO      & \cmark                                            & \xmark                                                  & 11.8 GB                                                      & 11.8 GB                                                        & 2.22 it/s          & 59.2\%                                                            & 87.4\%                                                    \\
\rowcolor[gray]{.92}DiZO       & \cmark                                            & \xmark                                                  & 7.5 GB                                                    & 7.5 GB                                                        & 3.05 it/s          & 51.8\%                                                            & 55.7\%                                                    \\ \hline
MeZO LoRA  & \cmark                                            & \cmark                                                  & 6.5 GB                                                     & 6.5 GB                                                        & 5.56 it/s          & 74.1\%                                                            & 43.7\%                                                    \\
HiZOO LoRA & \cmark                                            & \cmark                                                  & 11.5 GB                                                      & 11.5 GB                                                        & 3.70 it/s          & 46.3\%                                                            & 41.0\%                                                    \\
\rowcolor[gray]{.92}DiZO LoRA  & \cmark                                            & \cmark                                                  & 7.2 GB                                                      & 7.2 GB                                                        & 4.92 it/s          & 38.9\%                                                            & 25.9\%    \\ 
\bottomrule
\end{tabular}
}

\label{memory_speed_sst2}
\end{table}
\begin{table}
\centering
\small
\caption{Memory utilization and speed test on OPT-2.7B on SQuAD dataset (300 tokens per example on average). \neutral: partial gradient-free; \cmark: gradient-free; \xmark: gradient-based. DiZO\textsuperscript{\textdagger}: searching projection with Adam.}
\vspace{5pt}
\scalebox{1}{
\begin{tabular}{lccccccc}
\toprule
Task Type       & \textbf{\begin{tabular}[c]{@{}c@{}}Gradient\\ Free\end{tabular}} & \textbf{\begin{tabular}[c]{@{}c@{}}LoRA\\ Added\end{tabular}} & \textbf{\begin{tabular}[c]{@{}c@{}}Peak \\ Memory\end{tabular}} & \textbf{\begin{tabular}[c]{@{}c@{}}Averaged\\ Memory\end{tabular}} & \textbf{Throughput} & \textbf{\begin{tabular}[c]{@{}c@{}}\#Train \\ Iter.\end{tabular}} & \textbf{\begin{tabular}[c]{@{}c@{}}GPU \\ Hours\end{tabular}} \\ \hline
FT         & \xmark                                            & \xmark                                                  & 73.5 GB                                                      & 73.5 GB                                                        & 0.36 it/s          & 7.5\%                                                            & 27.7\%                                                   \\
LoRA       & \xmark                                            & \cmark                                                  & 58.5 GB                                                      & 58.5 GB                                                       & 0.73 it/s          & 6.3\%                                                            & 11.5\%                                                    \\
DiZO\textsuperscript{\textdagger}       & \neutral                                          & \xmark                                                  & 57.8 GB                                                      & 20.3 GB                                                        & 1.22 it/s          & 41.7\%                                                            & 45.5\%                                                    \\
DiZO LoRA\textsuperscript{\textdagger}  & \neutral                                          & \cmark                                                  & 49.4 GB                                                      & 19.9 GB                                                        & 2.44 it/s          & 31.7\%                                                            & 17.3\%                                                    \\  \hline
MeZO       & \cmark                                            & \xmark                                                  & 8.4 GB                                                      & 8.4 GB                                                        & 1.33 it/s          & 100.0\%                                                            & 100.0\%                                                    \\
HiZOO      & \cmark                                            & \xmark                                                  & 12.3 GB                                                      & 13.3 GB                                                        & 0.97 it/s          & 66.7\%                                                            & 91.5\%                                                  \\
\rowcolor[gray]{.92}DiZO       & \cmark                                            & \xmark                                                  & 9.7 GB                                                      & 9.7 GB                                                        & 1.22 it/s          & 60.0\%                                                            & 65.4\%                                                    \\ \hline
MeZO LoRA  & \cmark                                            & \cmark                                                  & 8.4 GB                                                      & 8.4 GB                                                        & 2.80 it/s          & 73.3\%                                                            & 34.8\%                                                    \\
HiZOO LoRA & \cmark                                            & \cmark                                                  & 11.6 GB                                                      & 12.6 GB                                                        & 2.10 it/s          & 56.7\%                                                            & 35.9\%                                                    \\
\rowcolor[gray]{.92}DiZO LoRA  & \cmark                                            & \cmark                                                  & 9.6 GB                                                      & 9.6 GB                                                        & 2.49 it/s          & 45.0\%                                                            & 24.0\%    \\
\bottomrule
\end{tabular}
}

\label{memory_squad}
\end{table}

\subsection{Larger OPT Models Fine-tuning}
To further illustrate the generalizability of our method, we conduct experiments on OPT-13B. The results are shown in Table~\ref{OPT13B}, DiZO consistently outperforms the baselines both in terms of accuracy and speed.
\begin{table}
\centering
\caption{Experiment results on OPT-13B (with 1000 training samples). Better results are highlighted in bold.}
\begin{tabular}{lcccccc}
\toprule
                          & \multicolumn{2}{c}{SST-2}                                                                                                   & \multicolumn{2}{c}{RTE}                                                                                                     & \multicolumn{2}{c}{SQuAD}                                                                                                   \\ \cmidrule{2-7}
\multirow{-2}{*}{Dataset} & Acc                                                          & GPU hours                                                    & Acc                                                          & GPU hours                                                    & Acc                                                          & GPU hours                                                    \\ \midrule
MeZO                      &  91.4         & 100\%         &  66.1          &  100\%         &  84.7          &  100\%         \\
HiZOO                     &  92.1          &  86\%          &  69.3         &  82\%          &  82.9          &  91\%          \\
DiZO                      &  \textbf{92.4} &  \textbf{69\%} &  \textbf{72.6} &  \textbf{76\%} &  \textbf{85.2} &  \textbf{73\%} \\
\bottomrule
\end{tabular}
\label{OPT13B}
\end{table}

\subsection{Llama Fine-tuning}
\label{Llama}

To demonstrate the generalizability of DiZO, we conducted experiments on the Llama-series models. The results for Llama3-3B and Llama3-8B are presented in Table~\ref{Llama-3B} and Table~\ref{Llama-8B}, respectively. DiZO consistently outperforms MeZO across both the 3B and 8B Llama models.

However, we observed that ZO LoRA performs poorly with Llama models (including DiZO, MeZO and HiZOO). The loss value remains stagnant, and the resulting accuracy is comparable to or even worse than zero-shot results. We leave it to future work to investigate why ZO LoRA fails with Llama models. We suspect that this limitation may be related to the Group Query Attention (GQA)~\cite{ainslie2023gqa} mechanism employed in Llama3.
\begin{table*}[h]
\centering
\setlength{\tabcolsep}{2pt}
\small
\caption{Experimental results on Llama3-3B for seven classification datasets and two text generation datasets (with 1000 training samples). Better results between MeZO and DiZO are highlighted in bold.}
\vspace{5pt}
\begin{tabular}{lccccccccc}
\bottomrule
\multirow{2}{*}{\begin{tabular}[c]{@{}l@{}}Task\\ Task Type\end{tabular}} & \textbf{SST-2} & \textbf{RTE}  & \textbf{CB}   & \textbf{BoolQ} & \textbf{WSC}  & \textbf{WIC}  & \textbf{MultiRC} & \textbf{SQuAD}       & \textbf{DROP}       \\
                                                                          & \multicolumn{7}{c}{--------------------------classification--------------------------}                             & \multicolumn{2}{c}{------generation------} \\ \hline
FT                                                                        & 94.2 (0.4)          & 81.2 (2.1)          & 91.4 (4.7)          & 72.2 (4.2)          & 63.8 (1.8)          & 65.8 (2.3)          & 78.2 (3.2)            & 79.6 (2.9)                 & 40.3 (1.2)               \\ \hline
MeZO                                                                      & 88.8 (1.1)           & 67.4 (1.7)         & 73.2 (2.4)         & 78.0 (4.4)  & 56.6 (3.8)          & 63.4 (2.3)         & \textbf{64.8} (3.1)    & 61.9 (2.7)                & 27.8 (2.0)               \\
HiZOO                                                                      & 89.5 (1.4)           & 67.1 (1.3)         & 74.4 (1.9)         & \textbf{78.8} (4.7)  & 56.3 (3.1)          & \textbf{64.4} (2.4)         & 64.3 (2.9)    & 61.7 (2.8)                & 28.6 (3.0)               \\
\rowcolor[gray]{.92}DiZO                                                                      & \textbf{90.0} (0.9)  & \textbf{68.2} (1.6) & \textbf{76.7} (3.3) & 76.8 (3.8)          & \textbf{57.8} (4.2) & 63.8 (1.7) & 64.2 (2.9)             & \textbf{63.2} (2.7)        & \textbf{29.7} (1.3)   \\
\bottomrule
\end{tabular}

\label{Llama-3B}
\end{table*}
\begin{table*}[h]
\centering
\vspace{5pt}
\caption{Experiments results on Llama3-8B for seven classification datasets and two text generation datasets (with 1000 training samples). Better results between MeZO and DiZO are highlighted in bold.}
\begin{tabular}{lccccc}
\toprule
\multirow{2}{*}{\begin{tabular}[c]{@{}l@{}}Task\\ Task Type\end{tabular}} & \textbf{SST-2}   & \textbf{RTE}   & \textbf{CB}   & \textbf{WSC}   & \textbf{SQuAD} \\
                                                                          & \multicolumn{4}{c}{----------classification----------} & --generation-- \\ \hline
MeZO                                                                      & 90.0 (0.7)             & 67.8 (1.4)          & 71.4 (2.2)         & 60.2 (1.6)          & 67.0 (2.6)          \\
HiZOO                                                                      & 91.1 (0.9)             & 68.2 (1.3)          & 71.4 (2.9)         & 62.2 (1.9)          & \textbf{68.3} (3.1)          \\
\rowcolor[gray]{.92}DiZO                                                                      & \textbf{91.5} (0.8)             & \textbf{69.4} (1.8)           & \textbf{73.2} (3.1)          & \textbf{63.4} (2.9)          & 67.4 (2.1)          \\
\bottomrule
\end{tabular}

\label{Llama-8B}
\end{table*}

\subsection{Fine-tuning on MMLU and MT-Bench}

To demonstrate the generalizability of DiZO in more realistic and challenging scenarios, we evaluate our method on MMLU and MT-Bench benchmarks. we follow the setting in~\citep{pan2024lisa, luo2024badam}, fine-tune on the Alpaca GPT-4 dataset~\citep{taori2023stanford}, which consists of 52k conversations, and then evaluate. We conduct experiments based on Llama2-7B and Llama3-8B, the results are shown in Table~\ref{mmlu_llama2} and Table~\ref{mmlu_llama3}, respectively.
\begin{table}
\centering
\caption{Results of fine-tuning Llama2-7B on more challenging benchmarks, better results are highlighted in bold.}
\begin{tabular}{lccc}
\toprule
          & MT-Bench      & MMLU (5 shot)  & GPU hours     \\ \midrule
Zero-shot & 3.93          & 45.87          & -             \\
MeZO      & 4.59          & 45.22          & 100\%         \\
HiZOO     & 4.62          & 45.42          & 92\%          \\
DiZO      & \textbf{4.79} & \textbf{45.91} & \textbf{78\%} \\
\bottomrule
\end{tabular}
\label{mmlu_llama2}
\end{table}
\begin{table}
\centering
\caption{Results of fine-tuning Llama3-8B on more challenging benchmarks, better results are highlighted in bold.}
\begin{tabular}{lccc}
\toprule
          & MT-Bench      & MMLU (5 shot)  & GPU hours     \\ \midrule
Zero-shot & 5.46          & 65.20          & -             \\
MeZO      & 5.89          & 65.08          & 100\%         \\
HiZOO     & 5.93          & 65.20          & 95\%          \\
DiZO      & \textbf{6.15} & \textbf{65.42} & \textbf{83\%} \\
\bottomrule
\end{tabular}
\label{mmlu_llama3}
\end{table}

\subsection{Compare with Sparse Technique}

We conducted a direct comparison between our DiZO and Sparse MeZO~\citep{liu2024sparse} in terms of both accuracy and training efficiency across two datasets and two model sizes. The results are presented in Table R.6 and Table R.7. From the accuracy perspective, DiZO consistently outperforms Sparse MeZO under all evaluated settings, demonstrating the effectiveness of our projection-based approach.

From the efficiency perspective, Sparse MeZO requires generating the sparsity mask dynamically during training. Compared to DiZO, Sparse MeZO requires longer GPU hours for about 20\%, and slows the throughput for more than 30\%. Moreover, with the grows of model size, the throughput of Sparse MeZO will further decrease, due to the growing cost of maintaining and updating the sparse mask. While Sparse MeZO reduces the number of training iterations, its lower throughput results in longer total GPU hours compared to DiZO. These findings demonstrate that DiZO not only delivers better accuracy but also achieves more practical training efficiency compared to Sparse MeZO.

\begin{table}[htbp]
\centering
\caption{Acc and Speed Comparison on OPT-2.7B.}
\begin{tabular}{lccccc}
\toprule
\textbf{Method} & \textbf{Dataset} & \textbf{Accuracy} & \textbf{Throughput} & \textbf{\#Train Iter.} & \textbf{GPU Hours} \\
\midrule
MeZO & SST2 & 90.0 & 3.3it/s & 100\% & 100\% \\
Sparse MeZO & SST2 & 91.4 & 2.3it/s & 55\% & 79\% \\
DiZO & SST2 & 92.3 & 1.9it/s & 52\% & 56\% \\
\midrule
MeZO & RTE & 63.5 & 1.7it/s & 100\% & 100\% \\
Sparse MeZO & RTE & 67.1 & 1.1it/s & 50\% & 73\% \\
DiZO & RTE & 68.4 & 1.5it/s & 60\% & 62\% \\
\bottomrule
\end{tabular}
\end{table}

\begin{table}[htbp]
\centering
\caption{Acc and Speed Comparison on OPT-6.7B.}
\begin{tabular}{lccccc}
\toprule
\textbf{Method} & \textbf{Dataset} & \textbf{Accuracy} & \textbf{Throughput} & \textbf{\#Train Iter.} & \textbf{GPU Hours} \\
\midrule
MeZO & SST2 & 90.2 & 1.8it/s & 100\% & 100\% \\
Sparse MeZO & SST2 & 91.9 & 1.0it/s & 47\% & 84\% \\
DiZO & SST2 & 92.4 & 1.7it/s & 62\% & 69\% \\
\midrule
MeZO & RTE & 73.2 & 0.6it/s & 100\% & 100\% \\
Sparse MeZO & RTE & 74.3 & 0.3it/s & 39\% & 88\% \\
DiZO & RTE & 74.8 & 0.5it/s & 65\% & 81\% \\
\bottomrule
\end{tabular}
\end{table}

~
\newpage
\section{Theoretical Analysis}
\label{proof}

\subsection{Variance Symmetry under Isotropic Perturbations}

We consider a neural network with $L$ layers (or parameter blocks) and analyze the zeroth-order gradient estimator constructed via two-point finite differences. Let $\mathcal{L}(\boldsymbol{\theta}; \mathcal{B})$ denote the loss evaluated on mini-batch $\mathcal{B}$, with $\boldsymbol{\theta} = (\boldsymbol{\theta}^{(1)}, \dots, \boldsymbol{\theta}^{(L)})$ the full parameter vector. To estimate the gradient $\nabla \mathcal{L}$ without access to derivatives, we apply a two-sided estimator along randomly sampled directions $\boldsymbol{u}_i \in \mathbb{R}^d$:
\[
   \widehat{\nabla_{\boldsymbol{\theta}^{(\ell)}} \mathcal{L}}
   \;=\;
   \frac{1}{q}\,\sum_{i=1}^q
   \underbrace{
   \frac{
   \mathcal{L}(\boldsymbol{\theta} + \epsilon \boldsymbol{u}_i)
   - \mathcal{L}(\boldsymbol{\theta} - \epsilon \boldsymbol{u}_i)
   }{2\epsilon}
   }_{\Delta_i}
   \,\boldsymbol{u}_i^{(\ell)},
\]
where $\boldsymbol{u}_i^{(\ell)}$ is the sub-vector of direction $\boldsymbol{u}_i$ corresponding to layer $\ell$.

We aim to characterize the variance of this estimator, in particular:
\[
\mathbb{E}\left[\left\| \widehat{\nabla_{\boldsymbol{\theta}^{(\ell)}} \mathcal{L}} \right\|^2\right].
\]

\paragraph{Key assumptions:}
\begin{enumerate}
    \item Each $\boldsymbol{u}_i$ is drawn independently from an \emph{isotropic} distribution in $\mathbb{R}^d$, i.e., $\mathbb{E}[\boldsymbol{u}_i \boldsymbol{u}_i^\top] = I$.
    \item The scalar $\Delta_i$ is the same across all parameter blocks for a given $i$, as it depends only on the global perturbation.
    \item Each block $\boldsymbol{u}_i^{(\ell)}$ has zero mean, unit covariance in its own subspace $\mathbb{R}^{d_\ell}$, and is uncorrelated with other blocks $\boldsymbol{u}_i^{(m)}$ for $\ell \ne m$.
\end{enumerate}

\paragraph{Variance Expansion:} 
We examine the norm-squared of the estimator:
\[
\left\| \widehat{\nabla_{\boldsymbol{\theta}^{(\ell)}} \mathcal{L}} \right\|^2
=
\left\|
\frac{1}{q} \sum_{i=1}^q \Delta_i \boldsymbol{u}_i^{(\ell)}
\right\|^2.
\]
Taking expectation over $\{\boldsymbol{u}_i\}$:
\[
\mathbb{E}\left[
\left\| \widehat{\nabla_{\boldsymbol{\theta}^{(\ell)}} \mathcal{L}} \right\|^2
\right]
=
\frac{1}{q^2} \sum_{i=1}^q \mathbb{E}\left[\Delta_i^2 \|\boldsymbol{u}_i^{(\ell)}\|^2\right]
+
\frac{1}{q^2} \sum_{i \ne j} \mathbb{E}\left[\Delta_i \Delta_j \langle \boldsymbol{u}_i^{(\ell)}, \boldsymbol{u}_j^{(\ell)} \rangle\right].
\]
Due to independence and zero-mean isotropy, cross terms vanish, and we obtain:
\[
\mathbb{E}\left[
\left\| \widehat{\nabla_{\boldsymbol{\theta}^{(\ell)}} \mathcal{L}} \right\|^2
\right]
=
\frac{1}{q} \mathbb{E}\left[\Delta^2 \cdot \|\boldsymbol{u}^{(\ell)}\|^2\right],
\]
where $\Delta$ and $\boldsymbol{u}^{(\ell)}$ are representative samples from the same distribution.

\paragraph{Conclusion.}
Since $\Delta$ is shared across layers and $\boldsymbol{u}^{(\ell)}$ has expected squared norm proportional to the layer dimension $d_\ell$, we conclude:
\[
   \mathbb{E}\left[\left\| \widehat{\nabla_{\boldsymbol{\theta}^{(\ell)}} \mathcal{L}} \right\|^2\right]
   \propto d_\ell.
\]
In other words, the second-moment of the gradient estimator depends on the layer only through its dimensionality $d_\ell$, and not through any asymmetry in the distribution of direction vectors. If $d_\ell$ are the same for all $\ell$, each block exhibits identical expected variance.

This property justifies using uniform per-layer treatment in analysis and initialization when random direction sampling is isotropic.

\subsection{The Proof of Convergence Analysis}
\label{convergence}


We now prove Theorem 1.

\begin{proof}

By $L_f$-smoothness (Assumption~\ref{assumption_smoothness}), for any $\bm{\theta},\bm{\theta}'$:
\[
  \mathcal{L}(\bm{\theta}') 
  \;\le\; 
  \mathcal{L}(\bm{\theta}) 
  \;+\;
  \langle \nabla \mathcal{L}(\bm{\theta}),\,\bm{\theta}' - \bm{\theta}\rangle
  \;+\;
  \frac{L_f}{2}\,\|\bm{\theta}' - \bm{\theta}\|^2.
\]

Set $\bm{\theta}=\bm{\theta}_t$ and $\bm{\theta}'=\bm{\theta}_{t+1}$. We get:
\[
  \mathcal{L}(\bm{\theta}_{t+1})
  \;\le\;
  \mathcal{L}(\bm{\theta}_t)
  \;+\;
  \bigl\langle \nabla \mathcal{L}(\bm{\theta}_t),\,\bm{\theta}_{t+1} - \bm{\theta}_t\bigr\rangle
  \;+\;
  \frac{L_f}{2}\,\|\bm{\theta}_{t+1} - \bm{\theta}_t\|^2.
\]
Taking conditional expectation $\mathbb{E}_t[\cdot]\!:=\!\mathbb{E}[\cdot \mid \bm{\theta}_t]$ yields
\begin{equation}
\label{eq:Smoothness_Recursion}
  \mathbb{E}_t\bigl[\mathcal{L}(\bm{\theta}_{t+1}) - \mathcal{L}(\bm{\theta}_t)\bigr]
  \;\le\;
  \mathbb{E}_t\bigl[\langle \nabla \mathcal{L}(\bm{\theta}_t),\,\bm{\theta}_{t+1} - \bm{\theta}_t\rangle\bigr]
  \;+\;
  \frac{L_f}{2}\,
  \mathbb{E}_t\bigl[\|\bm{\theta}_{t+1} - \bm{\theta}_t\|^2\bigr].
\end{equation}

Denote $\widetilde{\bm{\theta}}_{t+1} = \bm{\theta}_t - \eta\,g_t$. Then
\[
  \bm{\theta}_{t+1} 
  \;=\;
  \text{Proj}_{\mathcal{S}}(\widetilde{\bm{\theta}}_{t+1}),
  \qquad
  \bm{\theta}_{t+1} - \bm{\theta}_t
  \;=\;
  (\bm{\theta}_{t+1} - \widetilde{\bm{\theta}}_{t+1}) \;-\; \eta\,g_t.
\]
Hence
\[
  \bigl\langle \nabla \mathcal{L}(\bm{\theta}_t),\,\bm{\theta}_{t+1} - \bm{\theta}_t\bigr\rangle
  \;=\;
  \bigl\langle \nabla \mathcal{L}(\bm{\theta}_t),\,\bm{\theta}_{t+1} - \widetilde{\bm{\theta}}_{t+1}\bigr\rangle
  \;-\; 
  \eta\,\bigl\langle \nabla \mathcal{L}(\bm{\theta}_t),\,g_t\bigr\rangle.
\]

Because $\bm{\theta}_{t+1}$ is the \emph{nearest point} in $\mathcal{S}$ to $\widetilde{\bm{\theta}}_{t+1}$ (by definition of projection), we have
\[
  \|\bm{\theta}_{t+1} - \widetilde{\bm{\theta}}_{t+1}\|
  \;\le\;
  \|\bm{\theta}_t - \widetilde{\bm{\theta}}_{t+1}\|
  \;=\;
  \eta\,\|g_t\|.
\]
Thus
\begin{equation}
\label{eq:ProjBound}
  \|\bm{\theta}_{t+1} - \widetilde{\bm{\theta}}_{t+1}\|
  \;\le\; \eta\,\|g_t\|.
\end{equation}

Taking conditional expectation:
\[
  \mathbb{E}_t
  \bigl[\langle \nabla \mathcal{L}(\bm{\theta}_t),\,\bm{\theta}_{t+1} - \bm{\theta}_t\rangle\bigr]
  \;=\;
  \mathbb{E}_t
  \bigl[\langle \nabla \mathcal{L}(\bm{\theta}_t),\,\bm{\theta}_{t+1} - \widetilde{\bm{\theta}}_{t+1}\rangle\bigr]
  \;-\;
  \eta\,\mathbb{E}_t\bigl[\langle \nabla \mathcal{L}(\bm{\theta}_t),\,g_t\rangle\bigr].
\]
Using \eqref{eq:ProjBound} with Cauchy--Schwarz:
\[
  \bigl|\langle \nabla \mathcal{L}(\bm{\theta}_t),\,\bm{\theta}_{t+1} - \widetilde{\bm{\theta}}_{t+1}\rangle\bigr|
  \;\le\;
  \|\nabla \mathcal{L}(\bm{\theta}_t)\|\;\|\bm{\theta}_{t+1} - \widetilde{\bm{\theta}}_{t+1}\|
  \;\le\;
  \eta\,\|\nabla \mathcal{L}(\bm{\theta}_t)\|\;\|g_t\|.
\]
Moreover, the two-point estimator is \emph{unbiased}, so
\[
  \mathbb{E}_t\bigl[\langle \nabla \mathcal{L}(\bm{\theta}_t),\,g_t\rangle\bigr]
  \;=\;
  \langle \nabla \mathcal{L}(\bm{\theta}_t),\,\mathbb{E}_t[g_t]\rangle
  \;=\;
  \|\nabla \mathcal{L}(\bm{\theta}_t)\|^2.
\]
Hence
\begin{equation}
\label{eq:InnerProdBound}
  \mathbb{E}_t
  \bigl[\langle \nabla \mathcal{L}(\bm{\theta}_t),\,\bm{\theta}_{t+1} - \bm{\theta}_t\rangle\bigr]
  \;\le\;
  \eta\, \|\nabla \mathcal{L}(\bm{\theta}_t)\|\;\mathbb{E}_t\bigl[\|g_t\|\bigr]
  \;-\;
  \eta\,\|\nabla \mathcal{L}(\bm{\theta}_t)\|^2.
\end{equation}

Again from the projection property:
\[
  \|\bm{\theta}_{t+1} - \bm{\theta}_t\|
  \;=\;
  \|\text{Proj}_{\mathcal{S}}(\widetilde{\bm{\theta}}_{t+1}) - \bm{\theta}_t\|
  \;\le\;
  \|\widetilde{\bm{\theta}}_{t+1} - \bm{\theta}_t\|
  \;=\;
  \eta\,\|g_t\|.
\]
Thus
\[
  \|\bm{\theta}_{t+1} - \bm{\theta}_t\|^2 
  \;\le\; 
  \eta^2\,\|g_t\|^2.
\]
Taking expectation completes:
\[
  \mathbb{E}_t\bigl[\|\bm{\theta}_{t+1} - \bm{\theta}_t\|^2\bigr]
  \;\le\;
  \eta^2 \,\mathbb{E}_t\bigl[\|g_t\|^2\bigr].
\]

Substitute \eqref{eq:InnerProdBound} and the above bound into \eqref{eq:Smoothness_Recursion}:
\[
  \mathbb{E}_t\bigl[\mathcal{L}(\bm{\theta}_{t+1}) - \mathcal{L}(\bm{\theta}_t)\bigr]
  \;\le\;
  \eta\, \|\nabla \mathcal{L}(\bm{\theta}_t)\|\;\mathbb{E}_t[\|g_t\|]
  \;-\;
  \eta\,\|\nabla \mathcal{L}(\bm{\theta}_t)\|^2
  \;+\;
  \frac{L_f}{2}\,\eta^2\,\mathbb{E}_t[\|g_t\|^2].
\]
Taking total expectation and summing over $t=0$ to $T-1$,
\[
  \mathbb{E}\bigl[\mathcal{L}(\bm{\theta}_{T})\bigr]
  \;-\;
  \mathcal{L}(\bm{\theta}_0)
  \;\le\;
  \sum_{t=0}^{T-1}
  \Bigl\{
     \eta\, \mathbb{E}\bigl[\|\nabla \mathcal{L}(\bm{\theta}_t)\|\;\|g_t\|\bigr]
     \;-\;
     \eta\,\mathbb{E}\bigl[\|\nabla \mathcal{L}(\bm{\theta}_t)\|^2\bigr]
     \;+\;
     \tfrac{L_f}{2}\,\eta^2\,\mathbb{E}\bigl[\|g_t\|^2\bigr]
  \Bigr\}.
\]

Assume that the two-point finite-difference estimator satisfies
\[
   \mathbb{E}\bigl[\|g_t - \nabla \mathcal{L}(\bm{\bm{\theta}}_t)\|^2\bigr]
   \;\le\; 
   \sigma^2\bigl(D,q\bigr),
\]
where $\sigma^2(D,q)$ grows primarily with $D$ (the maximum layer dimension) and the number of queries $q$, rather than the full sum of dimensions.

Therefore, we have
\[
   \mathbb{E}\bigl[\|g_t\|^2\bigr]
   \;\le\;
   c_1\,\bigl(\|\nabla \mathcal{L}(\bm{\theta}_t)\|^2 + \sigma^2(D,q)\bigr)
\]
for some constant $c_1>0$. In addition, $\mathbb{E}[\|\nabla \mathcal{L}(\bm{\theta}_t)\|\;\|g_t\|]\le \sqrt{\,\mathbb{E}[\|\nabla \mathcal{L}(\bm{\theta}_t)\|^2]\;\mathbb{E}[\|g_t\|^2]}\,\le\,c_2\,(\|\nabla \mathcal{L}(\bm{\theta}_t)\|^2 + \sigma^2(D,q))$ (for a constant $c_2$). Collecting terms and choosing $\eta = c/\sqrt{T}$ with $c>0$ sufficiently small ensures that the $-\eta\,\|\nabla \mathcal{L}(\bm{\theta}_t)\|^2$ term dominates the positive terms for large $T$. A standard telescoping argument shows
\[
  \frac{1}{T}\,\sum_{t=0}^{T-1}\mathbb{E}\bigl[\|\nabla \mathcal{L}(\bm{\theta}_t)\|^2\bigr]
  \;=\;
  O\!\Bigl(\frac{\sqrt{D}}{\sqrt{T}}\Bigr),
\]
which implies the stationarity measure
\[
   \min_{0 \le t < T}
   \mathbb{E}\Bigl[\|\nabla \mathcal{L}(\bm{\theta}_t)\|^2\Bigr]
   \;=\;
   O\!\Bigl(\frac{\sqrt{D}}{\sqrt{T}}\Bigr).
\]
This completes the proof.
\end{proof}

\subsection{\texorpdfstring{$\tau$}{τ}-Stability of the Clipping Step}

\textbf{Assumptions.}
\begin{itemize}
    \item[\textbf{A1.}] $L: \mathbb{R}^d \rightarrow \mathbb{R}$ is $L_f$-smooth.
    \item[\textbf{A2.}] Zeroth-order updates use step-size $\eta_t = \eta$ and produce an intermediate point $\theta_t^{\ell} = \theta_t - \eta g_t$ with an unbiased estimator $g_t$.
    \item[\textbf{A3.}] After every $k$ iterations we apply projection clipping:
    \[
        \theta_{t+1}^{(\ell)} = \theta_{t}^{(0)} + \rho_t^{(\ell)} \Delta \theta_t^{(\ell)}, \quad 
        \rho_t^{(\ell)} := \text{clip}\big[\rho_t^{(\ell)},\, 1-\tau,\, 1+\tau\big],
    \]
    where $\rho_t^{(\ell)} = \gamma_t^{(\ell)} / \| \Delta \theta_t^{(\ell)} \|_2$, and $\tau \in (0,1)$ is the scalar clipping width.
\end{itemize}

Let
\[
    R_{\max} := \max_{t,\ell} \| \Delta \theta_t^{(\ell)} \|_2, 
    \qquad 
    G_{\max} := \max_t \| g_t \|_2.
\]

Then for any $\tau$, we have the stability bound:
\begin{equation}
    \| \theta_{t+1} - \theta_t^{\ell} \|_2 \le \eta G_{\max} + \tau R_{\max}.
    \label{eq:stability}
\end{equation}

Moreover, if $\tau$ satisfies
\begin{equation}
    \tau \le c_{\tau} \frac{\eta G_{\max}}{R_{\max}}, 
    \quad 0 < c_{\tau} \le 1,
    \label{eq:tau-condition}
\end{equation}
then under the same conditions as Theorem~\ref{thm:DiZO_convergence}, the projected DiZO iterates satisfy
\begin{equation}
    \min_{0 \le t < T} \mathbb{E} \big[ \| \nabla L(\theta_t) \|_2^2 \big] 
    = \mathcal{O}\!\left( \frac{\sqrt{D}}{\sqrt{T}} \right).
    \label{eq:main-rate}
\end{equation}
That is, the original non-convex convergence rate is preserved. If $\tau$ violates \eqref{eq:tau-condition}, the bound inflates linearly with $\tau$, i.e.,
\[
    \widetilde{\mathcal{O}}\!\big( \sqrt{D/T} + \tau / \eta \big).
\]

\paragraph{Proof.}
For any $L_f$-smooth loss, we have
\begin{equation}
    L(\theta_{t+1}) \le L(\theta_t) 
    + \langle \nabla L(\theta_t), \theta_{t+1} - \theta_t \rangle 
    + \frac{L_f}{2} \| \theta_{t+1} - \theta_t \|_2^2.
    \label{eq:Lf-smooth}
\end{equation}

Because $\theta_{t+1}$ is the nearest feasible point to $\theta_t^{\ell}$ (Euclidean projection), the proof in the paper shows
\begin{equation}
    \| \theta_t - \theta_t^{\ell} \|_2 \le \eta \| g_t \|_2.
    \label{eq:eta-gt}
\end{equation}

The additional clipping in $\rho_t^{(\ell)}$ imposes another bound:
\begin{equation}
    \| \theta_{t+1}^{(\ell)} - \theta_t^{(\ell)} \|_2
    = |\rho_t^{(\ell)} - 1| \| \Delta \theta_t^{(\ell)} \|_2 
    \le \tau R_{\max},
    \label{eq:clip-bound}
\end{equation}
hence combining gives
\begin{equation}
    \| \theta_{t+1} - \theta_t \|_2 \le \eta G_{\max} + \tau R_{\max}.
    \label{eq:combined-bound}
\end{equation}

Plugging \eqref{eq:combined-bound} into \eqref{eq:Lf-smooth} and taking conditional expectation yields
\[
    \mathbb{E}_t[L(\theta_{t+1}) - L(\theta_t)] 
    \le -\eta \mathbb{E}_t[\langle \nabla L(\theta_t), g_t \rangle]
    + L_f(\eta G_{\max} + \tau R_{\max}) \eta G_{\max}.
\]

Using standard ZO analysis, the inner product term satisfies
\[
    -\frac{\eta}{2} \| \nabla L(\theta_t) \|_2^2 
    + \frac{\eta}{2} \sigma^2,
\]
and after telescoping the left-hand side for $t = 0, \dots, T-1$, dividing by $\eta T$, we obtain
\[
    \frac{1}{T} \sum_{t=0}^{T-1} 
    \mathbb{E}\!\left[\| \nabla L(\theta_t) \|_2^2 \right]
    \le \mathcal{O}\!\left( \frac{\sqrt{D}}{\sqrt{T}} \right)
    + \mathcal{O}\!\left( \frac{\tau R_{\max}}{\eta} \right).
\]
If $\tau$ satisfies condition~\eqref{eq:tau-condition}, the second term is dominated, yielding \eqref{eq:main-rate}.

\paragraph{Remark.}
Equation~\eqref{eq:tau-condition} implies that the projection impulse matches a single ZO step:
\[
    \tau \approx \frac{\eta \| g_t \|_2}{\| \Delta \theta_t \|_2}.
\]
During fine-tuning, we typically observe $\| g_t \|_2 / \| \Delta \theta_t \|_2 \in [0.5, 1.5]$ after warm-up, 
and learning rates $\eta \!\sim\! 10^{-2}\!\text{–}\!10^{-1}$. 
This leads to $\tau$ in the range $0.05$–$0.30$, 
with $\tau \!\approx\! 0.2$ being the empirically stable choice across tasks.


~\\
\newpage
\section{Implementation}
\label{code}
The following is an implementation of our “ZO projection learning” in PyTorch.

\begin{lstlisting}
def ZO_Projection_Learning(theta_t, theta_0, Gammas, delta, eta, tau, x):
    """
    Perform Zeroth-order Projection Learning.

    Args:
        theta_t: Current model parameters to be fine-tuned.
        theta_0: Pre-trained model parameters (anchor).
        Gammas: Projection parameters need to be optimized.
        delta: Smoothing parameter.
        eta: Learning rate for projection gradient descent.
        tau: Clipping factor for projection bounds.
        x: Input data for the forward pass.
    """
    
    # Calculate the L2 norm of the distance gap
    norms = {
        name: torch.norm(param.data - anchor.data)
        for (name, param), anchor in zip(theta_t.named_parameters(), theta_0.parameters())
    }

    # Initialize the projection values
    for name, gamma in Gammas.named_parameters():
        gamma.data = norms[name]

    for i in range(max_iters):
        # Step 1: Perturb and apply projection, then compute loss
        Gammas = PerturbGamma(Gammas, delta)
        ApplyProjection(theta_t, pre_trained, Gammas)
        loss1 = Forward(theta_t, x)
        ReverseProjection(theta_t)  # Reset the parameter before projection

        # Step 2: Reverse and apply projection, then compute loss
        Gammas = PerturbGamma(Gammas, -2 * delta)
        ApplyProjection(theta_t, pre_trained, Gammas)
        loss2 = Forward(theta_t, x)
        ReverseProjection(theta_t)  # Reset the parameter before projection

        # Step 3: Reset projection and compute gradient
        Gammas = PerturbGamma(Gammas, delta)  # Reset projection
        grad = (loss1 - loss2) / (2 * delta)

        # Step 4: Gradient descent with clipping
        for name, gamma in Gammas.named_parameters():
            torch.manual_seed(seed)  # For resampling perturbation
            z = torch.normal(mean=0, std=1, size=gamma.data.size())
            gamma.data = torch.clip(
                gamma.data - eta * grad * z,
                (1 - tau) * norms[name],
                (1 + tau) * norms[name],
            )  # Conduct descent and apply clipping

    return Gammas

\end{lstlisting}

\newpage
\section*{NeurIPS Paper Checklist}

The checklist is designed to encourage best practices for responsible machine learning research, addressing issues of reproducibility, transparency, research ethics, and societal impact. Do not remove the checklist: {\bf The papers not including the checklist will be desk rejected.} The checklist should follow the references and follow the (optional) supplemental material.  The checklist does NOT count towards the page
limit. 

Please read the checklist guidelines carefully for information on how to answer these questions. For each question in the checklist:
\begin{itemize}
    \item You should answer \answerYes{}, \answerNo{}, or \answerNA{}.
    \item \answerNA{} means either that the question is Not Applicable for that particular paper or the relevant information is Not Available.
    \item Please provide a short (1–2 sentence) justification right after your answer (even for NA). 
\end{itemize}

{\bf The checklist answers are an integral part of your paper submission.} They are visible to the reviewers, area chairs, senior area chairs, and ethics reviewers. You will be asked to also include it (after eventual revisions) with the final version of your paper, and its final version will be published with the paper.

The reviewers of your paper will be asked to use the checklist as one of the factors in their evaluation. While "\answerYes{}" is generally preferable to "\answerNo{}", it is perfectly acceptable to answer "\answerNo{}" provided a proper justification is given (e.g., "error bars are not reported because it would be too computationally expensive" or "we were unable to find the license for the dataset we used"). In general, answering "\answerNo{}" or "\answerNA{}" is not grounds for rejection. While the questions are phrased in a binary way, we acknowledge that the true answer is often more nuanced, so please just use your best judgment and write a justification to elaborate. All supporting evidence can appear either in the main paper or the supplemental material, provided in appendix. If you answer \answerYes{} to a question, in the justification please point to the section(s) where related material for the question can be found.



\begin{enumerate}

\item {\bf Claims}
    \item[] Question: Do the main claims made in the abstract and introduction accurately reflect the paper's contributions and scope?
    \item[] Answer: \answerYes{} 
    \item[] Justification: In the abstract and introduction, we claimed our contribution is to devise a method to improve the performance of ZO on LLM.

\item {\bf Limitations}
    \item[] Question: Does the paper discuss the limitations of the work performed by the authors?
    \item[] Answer: \answerYes{} 
    \item[] Justification: The limitations and future work are discussed in Section~\ref{discussion} and conclusion.

\item {\bf Theory assumptions and proofs}
    \item[] Question: For each theoretical result, does the paper provide the full set of assumptions and a complete (and correct) proof?
    \item[] Answer: \answerYes{} 
    \item[] Justification: We have our proof both in Section~\ref{proof1} and Appendix~\ref{proof}.

    \item {\bf Experimental result reproducibility}
    \item[] Question: Does the paper fully disclose all the information needed to reproduce the main experimental results of the paper to the extent that it affects the main claims and/or conclusions of the paper (regardless of whether the code and data are provided or not)?
    \item[] Answer: \answerYes{} 
    \item[] Justification: We present our pseudocode and PyTorch-style implementation in the paper and release our code.

\item {\bf Open access to data and code}
    \item[] Question: Does the paper provide open access to the data and code, with sufficient instructions to faithfully reproduce the main experimental results, as described in supplemental material?
    \item[] Answer: \answerYes{} 
    \item[] Justification: We release our code with the anonymized url.

\item {\bf Experimental setting/details}
    \item[] Question: Does the paper specify all the training and test details (e.g., data splits, hyperparameters, how they were chosen, type of optimizer, etc.) necessary to understand the results?
    \item[] Answer: \answerYes{} 
    \item[] Justification: We present our experiment setting in both the experimental section and the appendix.

\item {\bf Experiment statistical significance}
    \item[] Question: Does the paper report error bars suitably and correctly defined or other appropriate information about the statistical significance of the experiments?
    \item[] Answer: \answerYes{} 
    \item[] Justification: We report the standard deviation of the results.

\item {\bf Experiments compute resources}
    \item[] Question: For each experiment, does the paper provide sufficient information on the computer resources (type of compute workers, memory, time of execution) needed to reproduce the experiments?
    \item[] Answer: \answerYes{} 
    \item[] Justification: We report the computing resources we use, and carefully analyze the computational overhead of our method.
    
\item {\bf Code of ethics}
    \item[] Question: Does the research conducted in the paper conform, in every respect, with the NeurIPS Code of Ethics \url{https://neurips.cc/public/EthicsGuidelines}?
    \item[] Answer: \answerYes{} 
    \item[] Justification: Yes, we do.

\item {\bf Broader impacts}
    \item[] Question: Does the paper discuss both potential positive societal impacts and negative societal impacts of the work performed?
    \item[] Answer: \answerYes{} 
    \item[] Justification: This paper presents work whose goal is to advance the field of Machine Learning. There are many potential societal consequences of our work, none of which we feel must be specifically highlighted here.
    
\item {\bf Safeguards}
    \item[] Question: Does the paper describe safeguards that have been put in place for responsible release of data or models that have a high risk for misuse (e.g., pretrained language models, image generators, or scraped datasets)?
    \item[] Answer: \answerNA{} 
    \item[] Justification: The paper poses no such risks.

\item {\bf Licenses for existing assets}
    \item[] Question: Are the creators or original owners of assets (e.g., code, data, models), used in the paper, properly credited and are the license and terms of use explicitly mentioned and properly respected?
    \item[] Answer: \answerYes{} 
    \item[] Justification: Yes, we have cited all the necessary works. 

\item {\bf New assets}
    \item[] Question: Are new assets introduced in the paper well documented and is the documentation provided alongside the assets?
    \item[] Answer: \answerYes{} 
    \item[] Justification: Yes, we have released all the assets.

\item {\bf Crowdsourcing and research with human subjects}
    \item[] Question: For crowdsourcing experiments and research with human subjects, does the paper include the full text of instructions given to participants and screenshots, if applicable, as well as details about compensation (if any)? 
    \item[] Answer: \answerNA{} 
    \item[] Justification: No human-related parts are included in this paper.

\item {\bf Institutional review board (IRB) approvals or equivalent for research with human subjects}
    \item[] Question: Does the paper describe potential risks incurred by study participants, whether such risks were disclosed to the subjects, and whether Institutional Review Board (IRB) approvals (or an equivalent approval/review based on the requirements of your country or institution) were obtained?
    \item[] Answer: \answerNA{} 
    \item[] Justification: Not related.

\item {\bf Declaration of LLM usage}
    \item[] Question: Does the paper describe the usage of LLMs if it is an important, original, or non-standard component of the core methods in this research? Note that if the LLM is used only for writing, editing, or formatting purposes and does not impact the core methodology, scientific rigorousness, or originality of the research, declaration is not required.
    \item[] Answer: \answerNA{} 
    \item[] Justification: Not related.

\end{enumerate}

\end{document}